\documentclass[lettersize,journal]{IEEEtran}
\usepackage{amsmath,amsfonts}
\usepackage{microtype}
\usepackage{graphicx}
\usepackage{booktabs} 

\usepackage{hyperref}
\usepackage{subcaption} 

\usepackage{mathtools}
\usepackage{amsthm}
\usepackage{multirow}
\usepackage{multicol}
\usepackage{bbding}
\usepackage{booktabs} 
\usepackage{algorithmic}
\usepackage{algorithm}
\usepackage[capitalize,noabbrev]{cleveref}
\usepackage{array}
\usepackage{cite}
\theoremstyle{plain}

\theoremstyle{definition}

\theoremstyle{remark}

\usepackage{wrapfig}
\usepackage{pifont}
\usepackage{amsfonts}  
\usepackage{nicefrac}  
\usepackage{xcolor} 
\usepackage{multirow}
\usepackage{multicol}
\usepackage{bbding}
\usepackage{booktabs} 
\usepackage[textsize=tiny]{todonotes}

\begin{document}
	\title{Harmonizing Generalization and Personalization \\ in Ring-topology Decentralized Federated Learning}
	\author{Shunxin~Guo, Jiaqi~Lv,~and~Xin Geng,~\IEEEmembership{Senior~Member,~IEEE}
		}
			%
			\markboth{Journal of \LaTeX\ Class Files,~Vol.~14, No.~8, August~2021}%
			{Shell \MakeLowercase{\textit{et al.}}: A Sample Article Using IEEEtran.cls for IEEE Journals}
			
			
			\maketitle
			
	\begin{abstract}
		We introduce Ring-topology Decentralized Federated Learning (RDFL) for distributed model training, aiming to avoid the inherent risks of centralized failure in server-based FL. However, RDFL faces the challenge of low information-sharing efficiency due to the point-to-point communication manner when handling inherent data heterogeneity. Existing studies to mitigate data heterogeneity focus on personalized optimization of models, ignoring that the lack of shared information constraints can lead to large differences among models, weakening the benefits of collaborative learning. To tackle these challenges, we propose a Divide-and-conquer RDFL framework (DRDFL) that uses a feature generation model to extract personalized information and invariant shared knowledge from the underlying data distribution, ensuring both effective personalization and strong generalization. Specifically, we design a \textit{PersonaNet} module that encourages class-specific feature representations to follow a Gaussian mixture distribution, facilitating the learning of discriminative latent representations tailored to local data distributions. Meanwhile, the \textit{Learngene} module is introduced to encapsulate shared knowledge through an adversarial classifier to align latent representations and extract globally invariant information. Extensive experiments demonstrate that DRDFL outperforms state-of-the-art methods in various data heterogeneity settings.
		

	\end{abstract}
	\section{introduction}
	Federated Learning (FL) is an emerging machine learning paradigm that enables multiple clients to collaboratively train a global model while keeping each client's private raw data stored locally~\cite{mcmahan2017communication,zhang2023trading,qi2023cross}. One of the main challenges in FL is data heterogeneity, which arises from significant distributional differences among clients~\cite{qu2022rethinking,  yang2024fedfed, li2024towards, qi2025cross}. Generally, effective solutions to mitigate this issue tend to focus on improving model generalization to better accommodate a wider range of clients, or enhancing model personalization to better adapt to local data distributions. Several classic Centralized Federated Learning (CFL) methods have been proposed, which regularize the model using a global model copy from the server~\cite{li2025personalized, li2021ditto, qu2023prevent} or decouple local models into shared parameters for server-side aggregation and partial personalization parameters for client-side customization~\cite{chen2021bridging, collins2021exploiting, oh2021fedbabu}. In addition, some approaches use both global and personalized classifiers downloaded from the server for prediction~\cite{chen2021bridging, zhang2023fedcp, xie2024perada}, aiming to mitigate inherent data shift issues due to data heterogeneity, thereby improving model performance. However, these methods generally rely on the presence of a central server, and this dependence on a centralized architecture can introduce problems such as single points of failure and server-client communication bottlenecks.
	\begin{figure*}[!tb]
		\begin{center}
			\vskip -0.1in
			\includegraphics[width=1\linewidth]{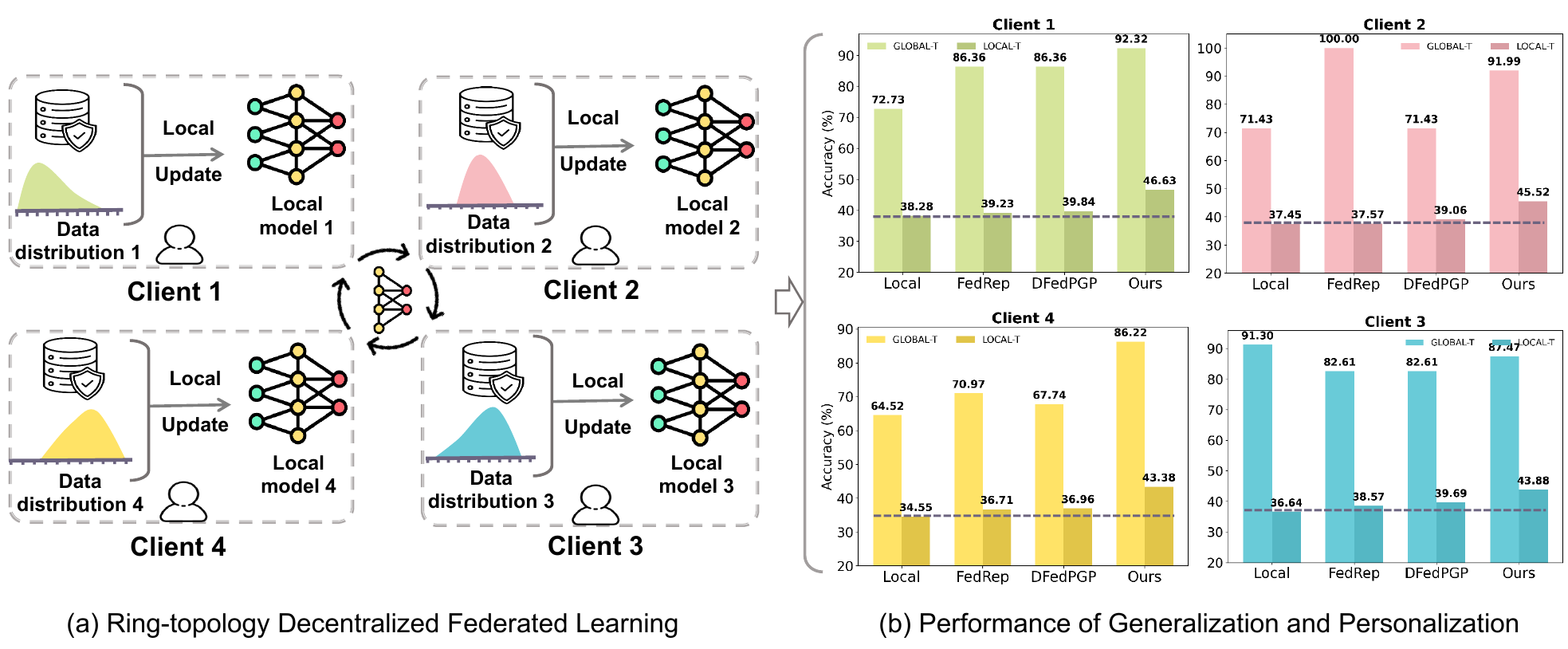}
			\caption{Illustration of the optimized learning mechanism in RDFL and comparison of the personalization and generalization performance of various advanced methods. (a) shows the schematic of the ring optimization approach, where client 1 receives the model from client 4, trains and updates it using its private dataset, and then transmits the locally updated model to client 2. The model is iteratively optimized over the ring topology. (b) presents the performance comparison of personalization (Local-T) and generalization (Global-T) on different methods across four clients with different data distributions. The results show that our proposed method effectively achieves both personalization and generalization goals in federated learning scenarios with data heterogeneity.}
			\label{fig:intro}
		\end{center}
		\vskip -0.2in
	\end{figure*}
	
	To avoid the limitations of server-based architecture, we focus on the Ring-topology Decentralized Federated Learning (RDFL) architecture that performs distributed model training~\cite{dai2022dispfl, li2023effectiveness, liu2024decentralized} using a client-directed communication manner, as shown in Figure~\ref{fig:intro}(a). However, its inherent architecture is challenging in addressing data heterogeneity, mainly due to the drift of data distribution across clients and the weakening of information sharing efficiency by point-to-point communication. The lack of effective global shared information for guidance leads to instability in the local model optimization process, exacerbating inconsistencies between client models. Existing DFL methods mainly focus on modeling individual user preferences through partial personalization or sparse masking of the model~\cite{kairouz2021advances, li2022learning, dai2022dispfl}, but these often overlook the information necessary for global generalization. The decoupled partial personalization models capture individual knowledge but fail to effectively leverage the remaining shared parameters to encode more generalizable knowledge~\cite{jeong2023personalized, li2024unleashing}. This is also confirmed by Figure~\ref{fig:intro} (b) with FL methods (local, FedRep~\cite{collins2021exploiting}, and DFedPGP~\cite{liu2024decentralized}) focusing on personalized optimization.

	Inspired by the recently proposed Learngene paradigm~\footnote{"Learngene" refers to a learning paradigm, while "\textit{Learngene}" denotes a model component encapsulating shared knowledge.}~\cite{wang2022learngene, wang2023learngene, xiainitializing}, which encapsulates shared knowledge within lightweight models to facilitate efficient task model initialization and adaptation to diverse data distributions, we propose a \textbf{D}ivide-and-conquer \textbf{R}ing-topology \textbf{D}ecentralized \textbf{F}ederated \textbf{L}earning (DRDFL) method to enhance communication efficiency and personalized learning capabilities.  
	To extract globally generalized knowledge that is independent of specific clients, we design a transferable \textit{Learngene} component that adversarial training using an adversarial classifier under a uniform class distribution constraint, allowing it to learn unbiased shared knowledge. Meanwhile, each client model adapts to its specific class distribution to address inconsistencies in data distributions across clients. To achieve this, we design a unique \textit{PersonaNet} component for each client, which undergoes personalized training based on
	the Gaussian distribution modeled by the mean and covariance of each class, and use the global Gaussian distribution of the corresponding class for auxiliary learning. The resulting personalized representation not only incorporates global information relevant to the target class but also enhances adaptation to heterogeneous data distributions. During inference, the invariant representation from \textit{Learngene} is combined with the personalized representation from \textit{PersonaNet} and fed into a decoder to reconstruct augmented data and generate the final prediction. Within the ring-topology framework, the \textit{Learngene} component and global implicit class representations undergo iterative optimization and sharing across clients, improving both generalization and personalization capabilities. Comprehensive experimental results demonstrate that the proposed DRDFL method effectively addresses data heterogeneity while improving communication efficiency, achieving a dual optimization of model generalization and personalization.

	To sum up, our contributions are as follows:
	\begin{itemize}
		\item 
		We analyze the challenges of data heterogeneity and inefficient information sharing faced by RDFL and argue that the essence is to harmonize the high generalization and personalized effectiveness of the model.
		\item We propose a novel method DRDFL that uses a feature generation process to extract both underlying shared knowledge across clients and personalized knowledge specific to each client.
		\item We empirically show the effectiveness of DRDFL, which outperforms eight SOTA methods by up to 3.28\% in local test accuracy with only \textbf{0.58 MB} of shared information among clients in each iteration. 
		
	\end{itemize}

\section{Related Work}
\subsection{Decentralized Federated Learning}
Federated learning (FL) has emerged as a promising research direction in the field of data privacy preservation~\cite{huang2022learn,tan2022fedproto,zhang2024fedtgp,chen2024recovering}. Due to the heterogeneity of computational and communication resources among clients, decentralized federated learning (DFL) has gained increasing attention in recent years. In this paradigm, clients connect solely with their neighbors through point-to-point communication to collaboratively train a consensus model. We investigate various approaches to personalization in DFL under multi-round local optimization. The classic DisPFL~\cite{dai2022dispfl} designs personalized models and pruning masks for each client, thereby accelerating personalized convergence. KD-PDFL~\cite{ liu2022decentralized} leverages knowledge distillation techniques to enable devices to distinguish statistical distances among local models. ARDM~\cite{sun2022decentralized} further proposes theoretical lower bounds on communication and local computation costs for personalized FL in a point-to-point communication framework. Recently, DFedPGP~\cite{liu2024decentralized} has adopted multi-step local updates and alternating optimization strategies to achieve superior convergence. These studies collectively advance the theoretical and practical development of personalized federated learning.
While showing promising results on personalization, the model exhibits inferior generalization performance, possibly due to the limited scalability of the input parameters. DRDFL can adapt intermediate features, enabling better generalization with greater adaptability to varying data distributions.


\subsection{Disentangled Representation Learning}
Disentangled Representation Learning has been proposed to address this limitation by learning models capable of identifying and disentangling the latent factors hidden within the observable data representations~\cite{wang2024disentangled,zhu2021commutative,guo2024dualvae}. %
Previous research has demonstrated the potential of using Variational auto-encoder (VAE)~\cite{kingma2013auto} to learn disentangled representations, which is a parametric model defined by $ p_{\theta}(\boldsymbol{x}|\mathbf{z}) $ and $ q_\phi(z|\boldsymbol{x}) $, employing the concept of variational inference to maximize the Evidence Lower Bound (ELBO), which is formulated as:
\begin{equation}
	\label{eq:ELBO}
	\log p(\boldsymbol{x}) \geq \mathbb{E}_{q_{\phi}(\mathbf{z} | \boldsymbol{x})}\left(\log p_{\theta}(\boldsymbol{x} | \mathbf{z})\right)-D_{\mathrm{KL}}\left(q_{\phi}(\mathbf{z} | \boldsymbol{x}) \| p(\mathbf{z})\right),
\end{equation}
where $ \mathbb{E}_{q_{\phi}(\mathbf{z} | \boldsymbol{x})}\left(\log p_{\theta}(\boldsymbol{x} | \mathbf{z})\right) $ represents the end-to-end reconstruction loss, and $ D_{\mathrm{KL}}\left(q_{\phi}(\mathbf{z} | \boldsymbol{x}) \| p(\mathbf{z})\right) $ is the KL divergence between the encoder's output distribution $ q_\phi(\mathbf{z}|\boldsymbol{x}) $ and the prior $ p(\mathbf{z}) $ that is usually modeled by standard normal distribution $ \mathcal{N}(0, \mathbf{I}) $. 
The further proposed CVAE~\cite{sohn2015learning} introduces label vectors as inputs to both the encoder and decoder, ensuring that the latent codes and generated images are conditioned on labels, which potentially mitigates latent collapse. Disentangled representation learning has been widely applied in CFL to address data heterogeneity~\cite{yan2023personalization}. Luo et al.~\cite{luo2022disentangled} proposed disentangled federated learning, leveraging invariant aggregation and diversity transfer to mitigate attribute bias. DisentAFL~\cite{chen2024disentanglement} adopts a two-stage disentanglement mechanism with a gating strategy to explicitly decompose the original asymmetric information-sharing scheme among clients into multiple independent and symmetric ones. Furthermore, FedST~\cite{wu2024spatio} achieves disentanglement and alignment of shared and personalized features across views and clients through orthogonal feature decoupling and regularization during both training and testing phases. 
Although their methods achieve remarkable performance, they rely on server-side model aggregation and associated proxy datasets, which are not aligned with the DFL paradigm. In contrast, our approach focuses on enhancing communication efficiency and optimizing consistency through direct interactions among clients.

\subsection{Learngene}
Learngene~\cite{lin2024linearly,xiainitializing,xia2024transformer,wangvision,li2024facilitating,feng2024wave,xie2024kind}, a novel machine learning paradigm inspired by biological genetics, has been proposed to distill large-scale ancestral models into generalized \textit{Learngene} that adaptively initialize models for various downstream tasks. Wang et al.~\cite{wang2022learngene} first introduced Learngene based on the gradient information of ancestral models and demonstrated its effectiveness in initializing new task models in open-world scenarios, thereby reflecting its strong generalization capability. To rapidly construct a diverse variety of networks with varying levels of complexity and performance trade-offs, the customized Learngene pool~\cite{shi2024building} methodology tailored to meet resource-constrained environments. Additionally, Feng et al.~\cite{feng2024transferring} further validated that transferring core knowledge through \textit{Learngene} is both sufficient and effective for neural networks. Inspired by this, we use \textit{Learngene} with generalized knowledge to transfer between clients to collaboratively learn others' knowledge, which provides a new way to simultaneously satisfy the conflicting goals of generalization and personalization.
\section{Preliminaries}
This section revisits the Ring-topology Distributed Federated Learning (RDFL) framework and analyzes the refinement of data heterogeneity issues. We explore the underlying motivations and discuss how to effectively address these challenges through divide-and-conquer collaboration.
\subsection{Problem Formulation}
In RDFL, client nodes are connected in a ring topology, where each client node communicates with only two neighboring clients. If a client node in the federated system fails, it will directly skip communication with the next client node, which has certain fault tolerance and flexibility. However, the point-to-point communication manner reduces the efficiency of global information sharing and amplifies the problem of data heterogeneity. Due to the inconsistency of data distribution among clients and the lack of effective global information to guide the optimization of local models, which in turn exacerbates the inconsistency between different client models. We revisit and refine the issue of data heterogeneity in the RDFL architecture as follows:

\textbf{Definition 3.1} (Feature Distribution Skew).\textit{ Let \( p_i(\boldsymbol{x}) \) and \( p_j(\boldsymbol{x}) \) represent the feature distributions for train sample \( i \) and test sample \( j \), respectively. The feature distribution of the training samples may be different from that of the test samples, but the class conditional distribution of the same class remains invariant, i.e.,}
\[
p_i(\boldsymbol{x}) \neq p_j(\boldsymbol{x}) \quad \text{but} \quad p_i(y|\boldsymbol{x}) = p_j(y|\boldsymbol{x}).
\]
\textbf{Definition 3.2 }(Label Distribution Skew). \textit{Let \( p_i(y) \) and \( p_j(y) \) denote the label distributions for client \( i \) and client \( j \), respectively. The label distributions across clients may differ, but the class-conditional feature distributions remain invariant, i.e.,}
\[
p_i(\boldsymbol{x}|y) = p_j(\boldsymbol{x}|y) \quad \text{but} \quad p_i(y) \neq p_j(y).
\]


\vspace{-3mm}
\subsection{Motivation}
Data heterogeneity is caused by the difference in data distribution among clients. The general approach to alleviate this issue is to personalize the local models by optimizing them according to their own data distributions. On the other hand, the inherently low information sharing efficiency of point-to-point communication in RDFL poses a challenge. Directly training with the model transmitted by neighboring client nodes, without accessing their local data, inevitably leads to performance degradation. In fact, the key to addressing data heterogeneity and improving the information sharing efficiency is to harmonize the model's generalization ability with its personalization needs. However, achieving these two conflicting goals simultaneously remains a highly challenging task.

We introduce a divide-and-conquer strategy to design independent and complementary dual modules to harmonize the generalization and personalization goals of the model. The Learngene~\cite{wang2022learngene,xiainitializing} paradigm involves encapsulating the common and invariant knowledge of large-scale models into lightweight information fragments, enabling subsequent task models to inherit and train on this knowledge, thereby achieving efficient knowledge transfer and utilization. Inspired by this concept, we design a \textit{Learngene} module that encapsulates global invariant knowledge for sharing across clients. This module effectively mitigates label distribution skew and enhances information-sharing efficiency. Meanwhile, each client locally trains a \textit{PersonaNet} module tailored to its specific data distribution to mitigate feature distribution skew.

\vspace{-3mm}
\subsection{Divide-and-Conquer Collaboration}
\textbf{Maximizing the learning of global knowledge is a reasonable way to address feature distribution skew.} Each client's local training and test datasets can be generated in different contexts/environments. For example, a client's training image samples may be primarily captured by a local camera, while the test images may come from the Internet and have different styles. From a contextual perspective, the target learning model must have a certain level of generalization capability to perform well in unknown and diverse contexts. We propose to train the \textit{PersonaNet} module based on the global class mean and variance derived from collaborative learning, allowing the capture of personalized information while mitigating feature distribution skew.

\textbf{Training with a uniform prior distribution is reasonable to address the label distribution skew.} Due to the inherent limitations of local clients, which are limited to their specific data subsets, they often fail to adequately represent the broader data distribution. Consequently, clients must collaborate to overcome the bottlenecks imposed by limited individual datasets. We emphasize the use of adversarial classifiers in training the \textit{Learngene} module within the RDFL system to adapt to a unified prior distribution $ p_u(y = k) = 1/K$, where $K$ represents the total number of classes. This ensures that cross-client collaboration is not affected by inconsistencies in class distributions, promoting the learning of a stable and invariant latent space, improving the generalization capability of the model.
\vspace{-2mm}
\section{Methodology}
\textbf{Notations.}
Consider a typical setting of RDFL with $ M $ clients, each client $ m $ has a dataset $ \mathcal{D}_{m}=\left\{\left(\boldsymbol{x}_{i}, y_{i}\right)\right\}_{i=1}^{\left|\mathcal{D}_{m}\right|} $, where $ y_{i} \in [1,K] $ and $K$ is the number of overall classes. The optimization problem that RDFL to solve can be formulated as:
\begin{equation}
\min _{\boldsymbol{w}_{m}} \mathcal{L}(\boldsymbol{w}_{m})=\frac{1}{\left|\mathcal{D}_{m}\right|} \sum_{i} \ell\left(\boldsymbol{x}_{i}, y_{i} ; \boldsymbol{w}_{m}\right),  
\end{equation}
where $ \boldsymbol{w}_m $ is the model parameter and $ \mathcal{L}(\boldsymbol{w}_{m})$ is the empirical risk computed from $m$-th client data $\mathcal{D}_{m}$; $ \ell $ is a loss function applied to each data instance.
\begin{figure}[!htb]
\centering
\includegraphics[width=1\linewidth]{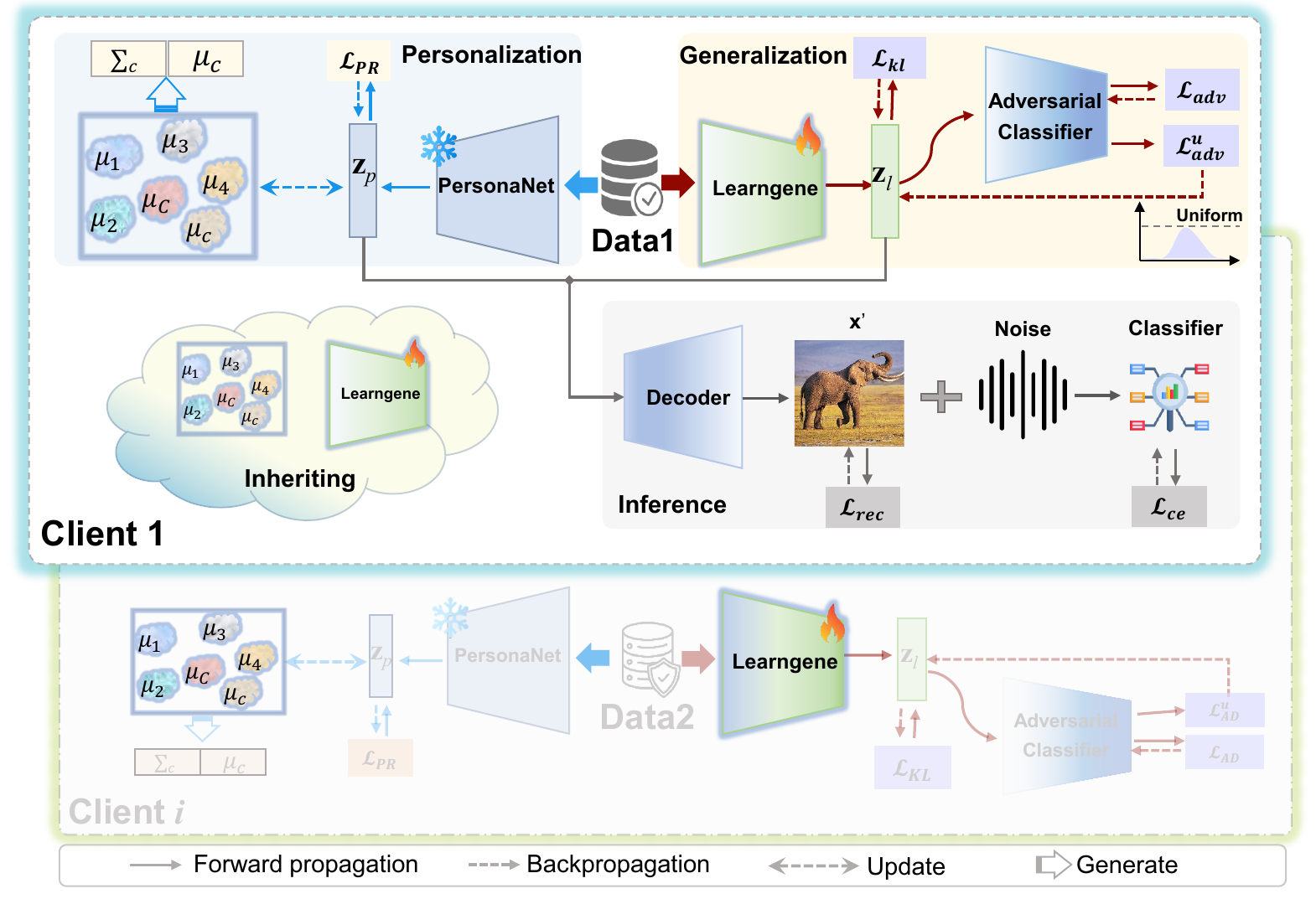}
\caption{
	Illustration of the DRDFL framework. It adopts a divide-and-conquer strategy, where \textit{PersonaNet} captures class-specific personalized representations via a Gaussian mixture distribution, enhancing adaptability to local data. Meanwhile, \textit{Learngene} employs adversarial alignment to ensure cross-client feature consistency and encapsulate globally invariant knowledge.  During inference, the decoder reconstructs inputs with noise for robustness, while iterative updates of global latent representations and \textit{Learngene} enhance information sharing across clients.}
\label{fig:framework}
\vskip-0.2in
\end{figure}

\textbf{Network architecture.} In the serverless DRDFL framework, the underlying goal of training a model with both personalized and generalized capabilities can be specifically described as: (1) identifying highly discriminative class-specific attributes to ensure accurate classification, and (2) mining class-independent common attributes to enhance the model's generalization ability. 
As shown in Figure~\ref{fig:framework}, by introducing a disentanglement mechanism based on the VAE, we design two complementary modules: a personalization module (\textit{PersonaNet}, parameterized by $ \psi_m $) to extract personalized representations, and a consensus module (\textit{Learngene}, parameterized by $ \phi $) to extract shared representations for collaborative learning among clients.
The decoder module $ p_{\theta_m} $ integrates the representations from both modules to reconstruct the input data, which is then passed to the prediction module $ f_{\omega_m}$ (parameterized by $ \omega $) for final decision making. Consequently, the local model $ \boldsymbol{w} $ can be decomposed as $ \boldsymbol{w}_m = [\psi_m, \phi, \theta_m, \omega_m]$ following a divide-and-conquer strategy, where $ \phi $ is used for shared aggregation across clients and other modules are private. For simplicity, we unify them without subscripts and focus on training a model with the dual objectives of generalization and personalization. 

The optimization of $[\psi, \phi, \theta]$ is achieved by maximizing the ELBO to provide a tight lower bound for the original $\log p(\boldsymbol{x})$ in Eq.~\ref{eq:ELBO}:     
\begin{equation}
\label{eq:ELBO_detail}
\begin{aligned}
	\max_{\psi, \phi, \theta} \, & \mathbb{E}_{\boldsymbol{x}}[\mathbb{E}_{q_{\psi}(\mathbf{z}_{p}, k | \boldsymbol{x}), q_{\phi}(\mathbf{z}_{l} | \boldsymbol{x})} \left[\log p_{\theta}(\boldsymbol{x} | \mathbf{z}_{p}, \mathbf{z}_{l}) \right] \\
	& - \underbrace{D_{\mathrm{KL}}\left(q_{\psi}(\mathbf{z}_{p}, k | \boldsymbol{x}) \| p(\mathbf{z}_{p}, k)\right)}_{\text{\textit{PersonaNet}}} \\
	& - \underbrace{D_{\mathrm{KL}}\left(q_{\phi}(\mathbf{z}_{l} | \boldsymbol{x}) \| p(\mathbf{z}_{l})\right)}_{\text{\textit{Learngene}}}],
\end{aligned}
\end{equation}
where \textbf{\textit{first}} term represents the negative reconstruction error. The \textbf{\textit{PersonaNet}} term enforces $ q_{\psi}(\mathbf{z}_p,k | \boldsymbol{x}) $ to align with the global class-specific prior Gaussian distribution, encouraging \textit{PersonaNet} to generate latent representations with strong class discriminability. Finally, the \textbf{\textit{Learngene}} term promotes the alignment of latent representations generated by \textit{Learngene} with the standard multivariate normal prior $ p(\mathbf{z}_l) $, thereby capturing shared information across classes.

\subsection{
Personalized \textit{PersonaNet} Training via Gaussian Mixture Distribution
}
The goal of the \textit{PersonaNet} module is to ensure model personalization to mitigate feature distribution skew. Based on the large-margin Gaussian mixture loss~\cite{wan2018rethinking, zheng2019disentangling}, we assume that the latent code \(\mathbf{z}_p\) learned from the training set follows a Gaussian mixture distribution expressed as:
\begin{equation}
p(\mathbf{z}_p) = \sum_{k} \mathcal{N}(\mathbf{z}_p; \boldsymbol{\mu}_k, \boldsymbol{\Sigma}_k) p(k),
\end{equation}
where $\boldsymbol{\mu}_k$ and \(\boldsymbol{\Sigma}_k\) represent the mean and covariance of class \(k\) in the feature space, and \(p(k)\) denotes the prior probability of class \(k\).
Under this assumption, we encourage \(\mathbf{z}_p\) to capture the necessary information related to the class label \(y\). Given a class label \(y \in [1, K]\), the conditional probability distribution of \(\mathbf{z}_p\) is defined as $p\left(\mathbf{z}_{p} | y\right) = \mathcal{N}\left(\mathbf{z}_{p}; \mu_{y}, \Sigma_{y}\right).$
Therefore, the corresponding posterior probability distribution can be represented:
\begin{equation}
p\left(y | \mathbf{z}_{p}\right)=\frac{\mathcal{N}\left(\mathbf{z}_{p} ; \mu_{y}, \Sigma_{y}\right) p\left(y\right)}{\sum_{k=1}^{K} \mathcal{N}\left(x_{i} ; \mu_{k}, \Sigma_{k}\right) p(k)}.
\end{equation}

Then maximizing the mutual information between \(\mathbf{z}_p\) and \(k\) is transformed into calculating the cross entropy between the posterior probability distribution and the one-hot encoded class label:
\begin{equation}
\begin{aligned}
\mathcal{L}_{cls} & =-\sum_{k=1}^{K} \mathbb{I}\left(y=k\right) \log q\left(k | \mathbf{z}_{p}\right) \\
& =- \log \frac{\mathcal{N}\left(\mathbf{z}_{p} ; \mu_{y}, \Sigma_{y}\right) p\left(y\right)}{\sum_{k=1}^{K} \mathcal{N}\left(x_{i} ; \mu_{k}, \Sigma_{k}\right) p(k)},
\end{aligned}
\end{equation}
where the indicator function \(  \mathbb{I}(\cdot) \) equals 1 if \(y\) is equal to \( k \), and 0 otherwise. Here, \( q(k | \mathbf{z}_p) \) refers to the auxiliary distribution introduced to approximate \( p(k | \mathbf{z}_p) \), since directly optimizing \( p(k | \mathbf{z}_p) \) is challenging in practice, as discussed in InfoGAN~\cite{chen2016infogan}.

Recall that in \textbf{\textit{PersonaNet}} term of Eq.~\ref{eq:ELBO_detail} the KL divergence between $q_{\psi}\left(\mathbf{z}_{p}, k|\boldsymbol{x}\right)$ and $p(\mathbf{z}_{p}, k)$ is minimized. 
If the covariance matrix of \(p(\mathbf{z}_p|y)\) tends to zero, the distribution tends to a degenerate Gaussian distribution, is expressed as $p(\mathbf{z}_p|y) \to \delta(\mathbf{z}_p - \boldsymbol{\mu}_y)$.
That is, all samples tend to the class mean $\boldsymbol{\mu}_y$. The KL divergence term degenerates into negative log-likelihood:
\begin{equation}
\mathcal{L}_{log} = -\log\mathcal{N}(\mathbf{z}_p; \boldsymbol{\mu}_y, \boldsymbol{\Sigma}_y),
\end{equation}
where $\mathbf{z}_p$ denotes the mean output from the \textit{PersonaNet}. The $\boldsymbol{\mu}_y$ and $\boldsymbol{\Sigma}_y$ 
dynamically updated using an Exponential Moving Average (EMA) strategy, i.e., $\boldsymbol{\mu}_y = \alpha \boldsymbol{\mu}_y + (1 - \alpha) \tilde{\boldsymbol{\mu}}_y, \boldsymbol{\Sigma}_y = \alpha \boldsymbol{\Sigma}_y + (1 - \alpha) \tilde{\boldsymbol{\Sigma}}_y$, where ($\tilde{\boldsymbol{\mu}}_y, \tilde{\boldsymbol{\Sigma}}_y$) denotes the inherited global class distribution. Finally the loss of \textit{PersonaNet} is defined as $\mathcal{L}_{PR}=\mathcal{L}_{cls}+ \mathcal{L}_{log}$.

\vspace{-1mm}
\subsection{Generalized \textit{Learngene} Training with Adversarial Classifier}
Intuitively, we aim to decompose the latent space $ \mathbf{z} $ such that $ \mathbf{z}_l $ adheres to a fixed prior distribution associated with knowledge shared across classes, independent of labels. This ensures that the resulting \textit{Learngene} encoding module possesses the advantage of being inheritable and transferable. Specially, the \textit{Learngene} term of in Eq.~\ref{eq:ELBO_detail} is implemented by minimizing the Kullback-Leibler divergence between $ q_{\phi}(\mathbf{z}_l|\boldsymbol{x}) $ and the prior $ p(\mathbf{z}_l) $ :
\begin{equation}
\mathcal{L}_{kl} = D_{\text{KL}}[q_{\phi}(\mathbf{z}_l | \boldsymbol{x}) \| p(\mathbf{z}_l)] = D_{\text{KL}}[\mathcal{N}(\boldsymbol{\mu},  \boldsymbol{\Sigma}) \| \mathcal{N}(0, I)],
\end{equation}
where $ q_{\phi}(\mathbf{z}_l|\boldsymbol{x}) $ is modeled as a Gaussian distribution with mean $ \boldsymbol{\mu} $ and diagonal covariance $\boldsymbol{\Sigma}$, both of which are the outputs of the \textit{Learngene}. 


To ensure that the \textit{Learngene} network exhibits generalization and that its output latent representations \(\mathbf{z}_l\) possess class-invariant properties, we design an adversarial classifier (parameterized by \(\vartheta\)) on \textit{Learngene} for adversarial training:
\begin{equation}
\mathcal{L}_{adv} = -\mathbb{E}_{\mathbf{z}_l \sim q_\phi(\mathbf{z}_l|\boldsymbol{x})} \log q_\vartheta(y|\mathbf{z}_l),
\end{equation}
where $ q_\vartheta(k| \mathbf{z}_l) $ represents the softmax probability output by the adversarial classifier.
Furthermore, we optimize \(\mathbf{z}_l\) with the goal of uniform class distribution to learn a class-indistinguishable latent representation, ensuring the class invariance of the representation:
\begin{equation}
\mathcal{L}_{\text{adv}}^u = - \mathbb{E}_{q_\phi(\mathbf{z}_l | \boldsymbol{x})} \left[ \frac{1}{C} \sum_{c=1}^{C} \log q_\vartheta(k| \mathbf{z}_l) \right].
\end{equation}
This strategy effectively avoids biased learning of specific categories within a single client, and can enhance the generalization training of the \textit{Learngene} module to achieve cross-client collaborative learning.
In summary, \textit{Learngene} captures generalized representations to achieve consistent optimization across clients, with the loss defined as:
$ \mathcal{L}_{GL} = \mathcal{L}_{kl}+\mathcal{L}_{\text{adv}}+\mathcal{L}_{\text{adv}}^u$.

\subsection{Robust Representation Learning via Noisy Reconstruction}

The latent representations produced by \textit{PersonaNet} and \textit{Learngene}, denoted as $ \mathbf{z}_p $ and $ \mathbf{z}_l $, are first concatenated and then fed into the decoder $ p_{\theta}(\boldsymbol{x}' \mid \mathbf{z}) $. The decoder parameters $ \theta$ are optimized by minimizing the $ L_2 $ reconstruction loss:
\begin{equation}
	\mathcal{L}_{\text{rec}} = \| \boldsymbol{x} - \boldsymbol{x}' \|_2^2.
\end{equation}

Although minimizing $\mathcal{L}_{rec}$ ensures that the generated sample $\boldsymbol{x}'$ closely approximates the original input $\boldsymbol{x}$, such high-fidelity reconstructions usually lack diversity. This may cause the learned \textit{Learngene} module to overfit to specific data instances, thereby weakening its generalization ability. To alleviate this problem, we inject Gaussian noise into the reconstructed samples during training to promote more robust and diverse gradient propagation during backpropagation:
$
\boldsymbol{x}_p = \boldsymbol{x}' + n$, where $n \sim \mathcal{N}(0, \sigma^2 \mathbf{I}).
$
This perturbation can reduce the risk of reconstructing the original data by encouraging \textit{Learngene} to capture more transferable and generalized representations.
%
%
%


Finally, a local classifier $f_\omega(\cdot)$ is trained on both the original and augmented data to simulate the label prediction process. The classification loss is defined as:
\begin{equation}
\begin{aligned}
	\mathcal{L}_{ce} &= \mathbb{E}_{(\boldsymbol{x}, y) \sim \mathcal{D}_{m}} \ell\left(f_\omega\left(\boldsymbol{x}\right), y\right) \\
	& + \mathbb{E}_{(\boldsymbol{x}_p, y) \sim P(\boldsymbol{x}_p, y)} \ell\left(f_\omega\left(\boldsymbol{x}_p\right), y\right),
\end{aligned}
\end{equation}
where $ \mathcal{D}_m $ is the local data distribution for client $ m $, $ P(\boldsymbol{x}_p, y) $ represents the distribution of the perturbed data and labels, and $ \ell(\cdot, \cdot) $ denotes the standard cross-entropy loss function.

Algorithm ~\ref{alg:DRDFL} describes the optimization process of $m$-th clinet local model in the DRDFL framework with a ring-topology decentralized training manner.
{\color{red}
\begin{algorithm}[!htb]
	\caption{DRDFL}
	\label{alg:DRDFL}
	\begin{algorithmic}[1]
		\STATE{\bfseries Input:}
		Total number of device $ M $, total number of communication rounds $ T $, local learning rate $\eta$, total number of classes $ K $, $m$-th client model $ \boldsymbol{w}_m = [\psi_m, \phi, \theta_m, \omega_m]$, pamrameter $\alpha$. 
		\STATE{\bfseries Output:} \textit{Learngene} $ \tilde{\phi}$ and $\{(\tilde{\boldsymbol{\mu}}_k , \tilde{\boldsymbol{\Sigma}}_k )\}_{k=1}^K$.
		\FOR{ $ t $ = 0 to $ T-1 $ }
		\FOR{ \textit{client} $ m $ \textit{in parallel}}
		\STATE Client $m$ receive neighbor's \textit{Learngene} $ \tilde{\phi} $ and $ \{(\tilde{\boldsymbol{\mu}}_k $,$\tilde{\boldsymbol{\Sigma}}_k )\}_{k=1}^K$.
		\STATE $ \boldsymbol{\mu}_k^{(m)} \leftarrow \alpha \boldsymbol{\mu}_k^{(m)} + (1 - \alpha) \tilde{\boldsymbol{\mu}}_k $ and $ \boldsymbol{\Sigma}_k^{(m)} \leftarrow \alpha \boldsymbol{\Sigma}_k^{(m)} + (1 - \alpha) \tilde{\boldsymbol{\Sigma}}_k $. 
		\STATE Set \textit{Learngene} $\phi \leftarrow \frac{\tilde{\phi} + \phi}{2}$  and sample a batch of local data $\xi_m$.
		\STATE \textbf{\textit{PersonaNet} Execution:}
		\STATE $ \psi_m \leftarrow-\eta\nabla_{\psi_m} \mathcal{L}_{PR}(\boldsymbol{\mu}_k^{(m)}, \boldsymbol{\Sigma}_k^{(m)}; \xi_m)$ .
		\STATE $ \boldsymbol{\mu}^{(m)}_{k},\boldsymbol{\Sigma}_{k}^{(m)} \leftarrow-\nabla_{\boldsymbol{\mu}_{k}^{(m)},\boldsymbol{\Sigma}_{k}^{(m)}} \mathcal{L}_{PR}$ .
		\STATE $ \mathbf{z}_{p} \leftarrow \psi_m (\xi_m)$ .
		\STATE \textbf{\textit{Learngene} Execution:}
		\STATE $ \phi \leftarrow-\eta\nabla_{\phi} \mathcal{L}_{GL}(\xi_m) $ .
		\STATE $ \mathbf{z}_{l} \leftarrow \phi (\xi_m) $ .
		\STATE \textbf{\textit{Decoder} and \textit{Classifier} Execution:}
		\STATE $ \xi_m^{\prime} \leftarrow \theta_m \left(\mathbf{z}_{p}, \mathbf{z}_{l}\right) $ .
		\STATE $ \theta_m \leftarrow -\eta\nabla_{\theta_m}\mathcal{L}_{rec}(\xi_m, \xi_m^{\prime}) $ .
		\STATE $ \omega_m \leftarrow -\eta\nabla_{\omega_m} \mathcal{L}_{ce}(\xi_m, \xi_m^{\prime}) $ .
		
		\ENDFOR
		\ENDFOR
		\STATE $\tilde{\phi}\leftarrow\phi, \{(\tilde{\boldsymbol{\mu}}_{k},\tilde{\boldsymbol{\Sigma}}_{k})\}_{k=1}^K\leftarrow\{(\boldsymbol{\mu}^{(m)}_{k},\boldsymbol{\Sigma}_{k}^{(m)})\}_{k=1}^K$.
	\end{algorithmic}
	\end{algorithm}	}
	\begin{table*}[!htbp]
\caption{Parameter-efficiency and averaged test accuracy across all clients' personalized models under the Dirichlet-based Non-IID setting. Note that \textbf{Bold}/\underline{Underline} fonts highlight the best/second-best approach.}
	\label{dir}
	\begin{center}
		\begin{small}
			\begin{sc}
				\scalebox{0.6}{
					\begin{tabular*}{28cm}{@{\extracolsep{\fill}}lcccccccccccccc}
							\toprule[1.2pt] 
							& &  & \multicolumn{4}{c}{\textbf{SVHN}} & \multicolumn{4}{c}{\textbf{CIFAR-10}} & \multicolumn{4}{c}{\textbf{CIFAR-100}} \\
							\cmidrule(lr){4-7} \cmidrule(lr){8-11} \cmidrule(lr){12-15}
							\textbf{Algorithm}&\textbf{Personalized }  & \textbf{\# Comm.}& \multicolumn{2}{c}{\textbf{ $ \beta $ = 0.1}}& \multicolumn{2}{c}{\textbf{ $ \beta $ = 0.4}}& \multicolumn{2}{c}{\textbf{ $ \beta $ = 0.1}}& \multicolumn{2}{c}{\textbf{ $ \beta $ = 0.4}}& \multicolumn{2}{c}{\textbf{ $ \beta $ = 0.1}}& \multicolumn{2}{c}{\textbf{ $ \beta $ = 0.4}}\\
							\cmidrule(lr){4-5} \cmidrule(lr){6-7} \cmidrule(lr){8-11} \cmidrule(lr){12-15}
							&\textbf{Params} &\textbf{Params} & Local-T & Global-T & Local-T & Global-T & Local-T & Global-T& Local-T & Global-T & Local-T & Global-T& Local-T & Global-T  \\
							\midrule
							Local   & Full model   & 0 \textit{M}      & 94.73 & 26.60 & 94.42 & 56.74 & 89.17   & 24.02 & 76.17 & 36.86 & 63.98  & 11.24 & 46.49 & 16.12 \\
							FedRep  & Output layer & 213.36\textit{M}   & 91.63 & 27.37 &\underline{94.70}&56.97 & 86.31   & 21.69 & 77.40 & 38.93 & 65.97  & 11.75 & 43.41 & 17.49 \\
							FedNova & Full model   & 213.46\textit{M}    & 95.70 & 31.70 &93.71& 56.74& \underline{90.85}   & 25.61 & 79.94 & 39.98 & 65.99  & 11.17 & 45.73 & 16.11 \\
							FedBN   & Batch norm.  &  213.06\textit{M}  &93.54 & \underline{32.98} & 94.32& \textbf{58.84} & 86.43 & 25.06&80.95&  39.92   &  63.75 & \textbf{15.01} & 36.01 & \textbf{19.09}\\
							FedFed  & Full model  & 454.58 \textit{M}    & 95.40 & 32.37 &93.65 &  56.68 & 91.30   & \underline{27.97} & \underline{82.70} & \underline{45.69} & 68.19  & 12.51 & \underline{48.67} & 16.54 \\
							\midrule
							DFedPGP & Output layer & 213.36\textit{M}    & \underline{95.98} & 30.22 & 92.62 & 54.65 & 87.62   & 25.46 & 78.06 & 42.72 & \underline{70.25}  & 11.28 & 43.15 & 16.77 \\
							
							FedCVAE& FULL MODEL & 63.44M & 76.81 & 14.12 & 78.82 & 41.47 & 78.17 & 14.62 & 78.82 & 41.43 & 58.13 & 8.12 & 40.09 & 11.53 \\
							DisPFL  & Masked model   &106.65\textit{M} &95.74&29.01&93.19&50.65& 89.43 &25.36& 79.54 & 38.83 & 58.39  & 9.95  & 47.56 & 15.59 \\
							DRDFL    & PersonaNet    & 0.58 \textit{M}  & \textbf{97.22}  &\textbf{33.09}&\textbf{94.73}& \underline{57.92}& \textbf{92.91} & \textbf{28.19} & \textbf{85.98} & \textbf{47.06} & \textbf{72.89}  & \underline{13.27} & \textbf{49.15} & \underline{17.60}\\
							\bottomrule
					\end{tabular*}}
				\end{sc}
			\end{small}
		\end{center}
		\vskip -0.1in
	\end{table*}
	\begin{table*}[ht]
		\caption{Parameter-efficiency and averaged test accuracy across all clients’ personalized models under the Shard-based Non-IID setting.}
			\label{shard}
			\label{sample-table}
			\begin{center}
				\begin{small}
					\begin{sc}
						\scalebox{0.6}{
								\begin{tabular}{lcccccccccccccc}
									\toprule[1.2pt] 
									& &  & \multicolumn{4}{c}{\textbf{SVHN}} & \multicolumn{4}{c}{\textbf{CIFAR-10}} & \multicolumn{4}{c}{\textbf{CIFAR-100}} \\
									\cmidrule(lr){4-7} \cmidrule(lr){8-11} \cmidrule(lr){12-15}
									\textbf{Algorithm}&\textbf{Personalized }  & \textbf{\# Comm.}& \multicolumn{2}{c}{\textbf{ $ s $ =4}}& \multicolumn{2}{c}{\textbf{ $ s $ = 5}}& \multicolumn{2}{c}{\textbf{ $ s $ = 4}}& \multicolumn{2}{c}{\textbf{ $ s $ = 5}}& \multicolumn{2}{c}{\textbf{ $ s $ = 20 }}& \multicolumn{2}{c}{\textbf{ $ s $ = 30}}\\
									\cmidrule(lr){4-5} \cmidrule(lr){6-7} \cmidrule(lr){8-11} \cmidrule(lr){12-15}
									&\textbf{Params} &\textbf{Params} & Local-T & Global-T & Local-T & Global-T & Local-T & Global-T& Local-T & Global-T & Local-T & Global-T& Local-T & Global-T  \\
									\midrule
									Local   & Full model   & 0\textit{M}      & 92.06 & 36.37 & 91.22 & 44.46 & 84.66 & 31.56 & 75.88 & 37.18 & 55.38 & 10.24 & 47.00 & 12.78 \\
									FedRep  & Output layer & 213.36\textit{M}   & 93.67 & 36.60 & 94.54 & 46.41 & 88.85 & 33.89 & 82.23 & 38.07 & 56.60 & 10.45 & 52.46 & 14.31 \\
									FedNova & Full model   & 213.45 \textit{M}   & 94.55 & 37.85 & \underline{95.30} & 46.60 & 88.12 & 33.51 & 82.75 & 39.57 & 57.62 & 11.12 & 54.29 & 13.84 \\
									FedBN   & Batch norm.  & 213.36 \textit{M}   & 92.98 & \underline{38.98} & 94.33 & \textbf{48.39} & \underline{90.51} & 34.41 & 83.58 & 42.80 & 59.99 & \textbf{13.62} & 55.88 & 14.87 \\
									FedFed  & Full model   & 454.58\textit{M}    & \underline{96.46} & 38.54 & 95.21 & 46.80 & 89.32 & \underline{35.63} & \underline{86.39} & \underline{42.99} & 67.68 & 12.72 & 53.81 & \underline{15.28} \\
									\midrule
									DFedPGP & Output layer & 213.36\textit{M}    & 91.94 & 37.06 & 92.36 & 45.79 & 87.11 & 32.20 & 80.54 & 38.64 & \underline{69.38} & 13.40 & \underline{58.30} & 13.68 \\
									FedCVAE & Full model   & 63.44  \textit{M}   & 86.21 & 34.09 & 78.22 & 39.33 & 70.86 & 26.98 & 75.03 & 37.11 & 63.81 & 11.98 & 52.60 & 14.90 \\
									DisPFL  & Full model   & 106.65\textit{M} & 94.36 & 37.62 & 95.29 & 46.76 & 87.05 & 33.70 & 80.95 & 39.60 & 60.16 & 10.89 & 54.02 & 13.55 \\
									DRDFL    & PersonaNet    & 0.58  \textit{M}    & \textbf{96.72} & \textbf{39.98} & \textbf{95.36 }& \underline{47.74} & \textbf{92.30} & \textbf{36.72} & \textbf{89.57} & \textbf{44.66 }& \textbf{71.24} & \underline{13.45} & \textbf{58.60} & \textbf{16.56}\\
									\bottomrule								\end{tabular}}
						\end{sc}
					\end{small}
				\end{center}
				\vskip -0.1in
			\end{table*}
			
			\section{Experiments}
			\subsection{Experiment Setup}
			\textbf{Implementation.}
			We implemented the proposed method and the considered baselines in PyTorch. The models are trained using ResNet18 in a simulated decentralized ring topology federated learning environment with multiple participating clients. By default, the number of clients is set to 20, the learning rate is set to 0.001, the number of global training rounds is set to 300, the number of local update epochs is set to 5, and the batch size is set to 64.
			
			\textbf{Evaluation metrics.} We report the mean test accuracy of personalized models for all clients. The evaluation is based on two primary metrics: Local-T (i.e., using the local test data corresponding to each client's class distribution) and Global-T (i.e., using the union of all clients' local test data). These metrics are used to assess the model's personalization performance (feature distribution skew) and generalization ability (label distribution skew).
			
			\textbf{Dataset and data partition.}
			We use 3 real-world datasets: SVHN~\cite{netzer2011reading}, CIFAR-10 and CIFAR-100~\cite{krizhevsky2009learning}, which are widely used for evaluating FL algorithms. 
			To comprehensively simulate real-world scenarios, we adopt two types of Non-IID data distributions:  
			\begin{itemize}
				\item Dirichlet-based Non-IID: This method partitions the training data according to a Dirichlet distribution Dir($ \beta $), while the corresponding test data for each client is generated from the same distribution~\cite{chen2021communication,dai2022dispfl}. We set $ \beta $ = \{0.1, 0.4\} for all benchmark datasets.
				\item Shard-based Non-IID: The dataset is grouped by class (each class corresponding to one shard), and the shards are then unevenly distributed across clients. The sample distribution received by each client differs, and the heterogeneity is typically controlled by adjusting the number of classes allocated to each client. Specifically, we set $ s = \{4, 5\} $ for the SVHN and CIFAR-10 datasets, and $ s = \{20, 30\} $ for the CIFAR-100 dataset.  
			\end{itemize}
			{\color{red}
				\begin{figure*}[ht]
					
					\begin{center}
						\begin{minipage}[b]{0.24\linewidth}
							\centering
							\includegraphics[width=\linewidth]{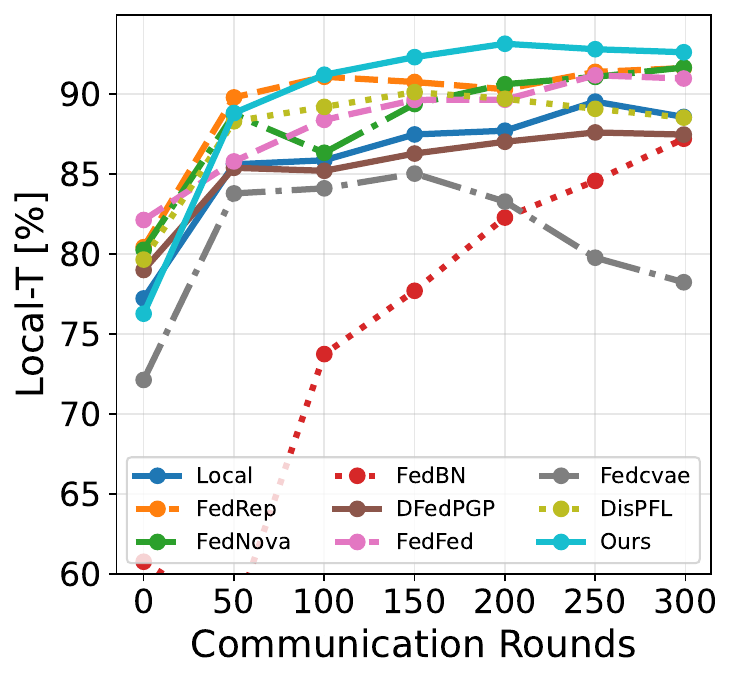}
							\subcaption{CIFAR-10 with $ \beta $ = 0.1}
							\label{fig:subplot1}
						\end{minipage}
						\begin{minipage}[b]{0.24\linewidth}
							\centering
							\includegraphics[width=\linewidth]{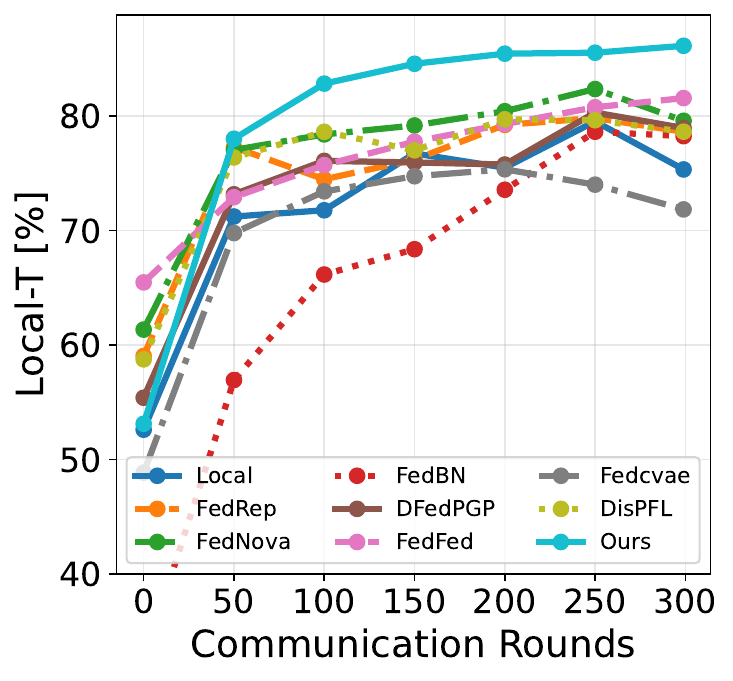}
							\subcaption{CIFAR-10 with $ \beta $ = 0.4}
							\label{fig:subplot2}
						\end{minipage}
						\begin{minipage}[b]{0.24\linewidth}
							\centering
							\includegraphics[width=\linewidth]{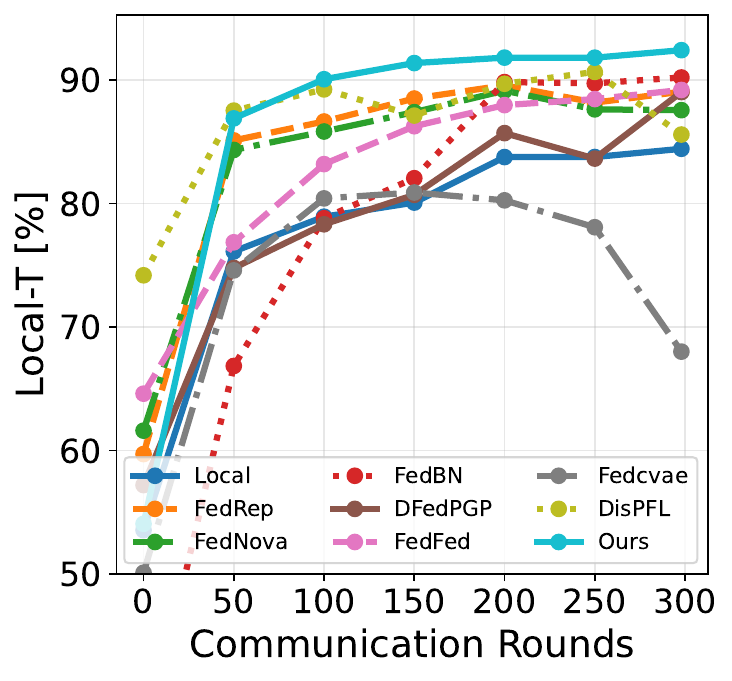}
							\subcaption{CIFAR-10 with $ s $ = 4}
							\label{fig:subplot3}
						\end{minipage}
						\begin{minipage}[b]{0.24\linewidth}
							\centering
							\includegraphics[width=\linewidth]{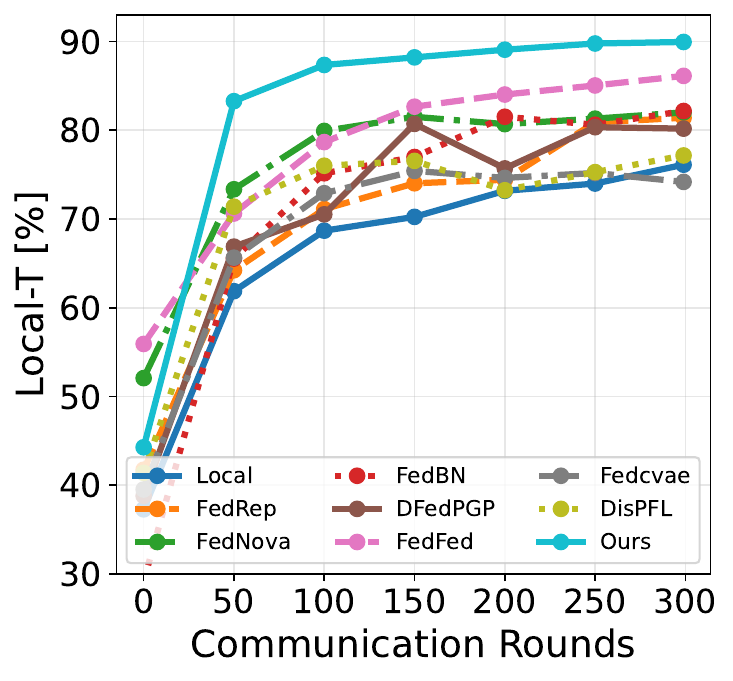}
							\subcaption{CIFAR-10 with $ s $ = 5}
							\label{fig:subplot4}
						\end{minipage}\vskip -0.1in
						\caption{Comparison of Local-T curves for different methods under various Non-IID settings on the CIFAR-10 dataset.}
						\label{CON_CIFAR10}
					\end{center}
					\vskip -0.3in
			\end{figure*}}
			
			\textbf{Baselines.}
			We selected a series of state-of-the-art federated learning algorithms for comparison, including \textit{Local}, which performs training locally without collaboration, and CFL methods designed to mitigate data heterogeneity, such as FedRep~\cite{collins2021exploiting}, FedNova~\cite{wang2020tackling}, FedBN~\cite{li2021fedbn}, and FedFed~\cite{yang2024fedfed}. Furthermore, DFL approaches, including DFedPGP~\cite{liu2024decentralized}, Fedcvae, and DisPFL~\cite{dai2022dispfl}, were used as baselines. All methods were evaluated using ResNet-18 and involved collaborative learning between all clients in the FL system. Detailed description is given in Appendix~\ref{data}.
			
			\subsection{Evaluation Results}
			\textbf{DRDFL is parameter-efficient.} 
			Tables~\ref{dir} and~\ref{shard} report the parameters of the personalized module trained on each client and the communication cost per round transmitted to the server under different settings.
			In terms of communication efficiency, DRDFL significantly outperforms most state-of-the-art CFL and DFL methods across different settings. By locally training two modules, \textit{Learngene} and Personanet, which address different tasks, and by leveraging \textit{Learngene} for communication, DRDFL achieves highly efficient and low-cost communication. Classic CFL methods, such as FedRep, FedNova, and FedBN, require approximately 213M parameters to be sent to the server for aggregation. In contrast, DRDFL only exchanges 0.58M parameters, including the \textit{Learngene} module and negligible class mean vectors, greatly reducing communication overhead. While FedFed requires the upload of the model and generated datasets, DFL-based methods, such as DispFL, still rely on the transmission of masked models and low-level model parameters in interactions, failing to match DRDFL's advantage in parameter efficiency.
			
			\textbf{DRDFL achieves competitive personalized performance and better generalization than baselines.} Tables~\ref{dir} and~\ref{shard} show that, compared to DFL methods on the same architecture, DRDFL delivers competitive personalization results with significantly fewer communication parameters. 
			In particular, in the Dirichlet-based $ \beta = 0.1 $ setting, DRDFL outperforms the state-of-the-art DFedPGP method ($ + $ 1.24\%, 5.29\%, 2.64\% on SVHN, CIFAR-10, and CIFAR-100). DRDFL also achieves comparable generalization performance to central server CFL methods on CIFAR-10, providing competitive results. While it performs slightly worse than the FedBN method ($ - $ 1.74\%, 0.17\% at $ \beta $ = 0.1, $ s $ = 20), which aggregates batch normalization models, on the CIFAR-100 dataset. However, DRDFL achieves the best performance that is higher than FedFed ($ + 1.28\%$) on CIFAR-100 with $ s = 30 $. These results highlight the effectiveness of \textit{Learngene} as a shared component for iterative optimization across clients, enabling generalized knowledge learning in distributed federated learning scenarios.
			
			
			\textbf{Convergence analysis.} We demonstrate the personalized performance of the model from a convergence perspective in Figure~\ref{CON_CIFAR10}, which shows the performance curves of the FL method across different partitioning schemes on the CIFAR-10 dataset over communication rounds. Our method uses a divide-and-conquer strategy to achieve the dual goals of model personalization and generalization without introducing convergence-related problems. Compared to other state-of-the-art methods, DRDFL achieves the best convergence speed under different partition settings and converges to higher personalized performance. In particular, it is more significant on shard-based Non-IID data partitions. The personalized performance of DRDFL is already higher than other methods at \textbf{Round} 50 and gradually increases in the subsequent training stages to reach a convergence state. This shows that our method is effective in achieving the personalized goal of the model.
			
			\textbf{Ablation study.}
			In Table~\ref{tab:le}, we present the impact of different components on the overall method, evaluated using the Local-T and Global-T metrics for the CIFAR-10 dataset with different data partitions. When the DRDFL method does not include the personalized \( \mathcal{L}_{PR} \) component, its performance on Local-T is significantly worse, while Global-T remains roughly unchanged. In contrast, omitting the \( \mathcal{L}_{GL} \) component, which controls for generalization invariant representations, slightly decreases Global-T performance. 
			This indicates that using different components to divide-and-conquer personalization and generalization is effective and indispensable to the overall model. Additional experimental results are presented in Appendix~\ref{aer}.
			\begin{table}[!tb]
				\caption{Ablation studies for DRDFL on the CIFAR-10 dataset.}
				\label{tab:le}
				\begin{center}
					\begin{small}
						\begin{sc}
							\scalebox{0.65}{
								\begin{tabular*}{13cm}{@{\extracolsep{\fill}}l|ll|cccc}
									\toprule[1.2pt]
									&&  & \multicolumn{2}{c}{\textbf{$ s $ = 4}}&\multicolumn{2}{c}{\textbf{$\beta $ = 0.1}}  \\
									\multirow{-2}{*}{\textbf{Settings}} &\multirow{-2}{*}{ $ \mathcal{L}_{PR} $ } &\multirow{-2}{*}{ $ \mathcal{L}_{GL} $ } &  Local-T & Global-T& Local-T & Global-T \\
									\midrule
									DRDFL w/o $ \mathcal{L}_{PR} $ &\XSolidBrush & \Checkmark  & 90.26&36.42&90.01&27.84  \\
									DRDFL w/o $ \mathcal{L}_{GL} $ & \Checkmark & \XSolidBrush &91.39&34.32&92.63&26.84\\
									DRDFL &\Checkmark & \Checkmark&\textbf{92.30}&\textbf{36.72}&\textbf{92.91}&\textbf{28.19} \\
									\bottomrule[1.2pt]
							\end{tabular*}}
						\end{sc}
					\end{small}
				\end{center}
				\vskip -0.3in
			\end{table}

			
			\section{Conclusions}
			In this paper, we propose a divide-and-conquer approach based on a ring-based distributed federated learning architecture. By designing two learning modules, \textit{PersonaNet} and \textit{Learngene}, we decouple the goals of generalization and personalization. The former leverages Gaussian mixture learning to enhance class separability, while the latter uses adversarial learning to extract invariant representations, achieving a dual benefit of both generalization and personalization. We demonstrate the effectiveness of the proposed method across various datasets.
			\onecolumn
			\section{Appendix}
			\subsection{The ELBO of the log-likelihood objective}
			First, the \textit{PersonaNet} network outputs a class representation \( \mathbf{z}_p \sim p(\mathbf{z}_p, k) \), and the \textit{Learngene} outputs a cross-class independent representation \( \mathbf{z}_l \sim p(\mathbf{z}_l) \). Then, the decoder \( p_\theta(x | \mathbf{z}_p, \mathbf{z}_l) \) takes the combination of \( \mathbf{z}_p \) and \( \mathbf{z}_l \) as input and maps the latent representations to images. Therefore, we decompose the joint distribution \( p(x, \mathbf{z}_p, \mathbf{z}_l) \) as follows:
			\begin{equation}
				p\left(\boldsymbol{x}, \mathbf{z}_{p}, \mathbf{z}_{l}\right)=\sum_{k} p_{\theta}\left(\boldsymbol{x} | \mathbf{z}_{p}, \mathbf{z}_{l}\right) p\left(\mathbf{z}_{p}, k\right) p\left(\mathbf{z}_{l}\right)
			\end{equation}
			By using Jensens inequality, the log -likelihood $\log p(\boldsymbol{x})$  can be written as:
			\begin{equation}
				\begin{aligned}
					\log p(\boldsymbol{x})= & \log \iint p\left(\boldsymbol{x}, \mathbf{z}_{p}, \mathbf{z}_{l}\right) d \mathbf{z}_{p} d \mathbf{z}_{l} \\
					= & \log \iint \sum_{k} p_{\theta}\left(\boldsymbol{x} | \mathbf{z}_{p}, \mathbf{z}_{l}\right) p\left(\mathbf{z}_{p}, k\right) p\left(\mathbf{z}_{l}\right) d \mathbf{z}_{p} d \mathbf{z}_{l} \\
					= & \log \mathbb{E}_{q_{\psi}\left(\mathbf{z}_{p}, k | \boldsymbol{x}\right), q_{\phi}\left(\mathbf{z}_{l} | \boldsymbol{x}\right)} \frac{p_{\theta}\left(\boldsymbol{x} | \mathbf{z}_{p}, \mathbf{z}_{l}\right)  p\left(\mathbf{z}_{p}, k\right)p\left(\mathbf{z}_{l}\right)}{q_{\psi}\left(\mathbf{z}_{p}, k | \boldsymbol{x}\right) q_{\phi}\left(\mathbf{z}_{l} | \boldsymbol{x}\right)} \\
					\geq & \mathbb{E}_{q_{\psi}\left(\mathbf{z}_{p}, k | \boldsymbol{x}\right), q_{\phi}\left(\mathbf{z}_{l} | \boldsymbol{x}\right)}\left[\log \frac{p_{\theta}\left(\boldsymbol{x} | \mathbf{z}_{p}, \mathbf{z}_{l}\right) p\left(\mathbf{z}_{p}, k\right)p\left(\mathbf{z}_{l}\right) }{q_{\psi}\left(\mathbf{z}_{p}, k | \boldsymbol{x}\right) q_{\phi}\left(\mathbf{z}_{l} | \boldsymbol{x}\right)}\right] \\
					= & \mathbb{E}_{q_{\psi}\left(\mathbf{z}_{p}, k | \boldsymbol{x}\right), q_{\phi}\left(\mathbf{z}_{l} | \boldsymbol{x}\right)}\left[\log p_{\theta}\left(\boldsymbol{x} | \mathbf{z}_{p}, \mathbf{z}_{l}\right)\right] \\
					& +\mathbb{E}_{q_{\psi}\left(\mathbf{z}_{p}, k | \boldsymbol{x}\right), q_{\phi}\left(\mathbf{z}_{l} | \boldsymbol{x}\right)}\left[\log \frac{p\left(\mathbf{z}_{p}, k\right)}{q_{\psi}\left(\mathbf{z}_{p}, k| \boldsymbol{x}\right)}\right]\\
					& +\mathbb{E}_{q_{\psi}\left(\mathbf{z}_{p}, k | \boldsymbol{x}\right), q_{\phi}\left(\mathbf{z}_{l} | \boldsymbol{x}\right)}\left[\log \frac{p\left(\mathbf{z}_{l}\right)}{q_{\phi}\left(\mathbf{z}_{l} | \boldsymbol{x}\right)}\right] \\
					=&\mathbb{E}_{q_{\psi}\left(\mathbf{z}_{p}, k | \boldsymbol{x}\right), q_{\phi}\left(\mathbf{z}_{l} | \boldsymbol{x}\right)}\left[\log p_{\theta}\left(\boldsymbol{x} | \mathbf{z}_{p}, \mathbf{z}_{l}\right)\right] \\
					& -D_{\mathrm{KL}}\left(q_{\psi}\left(\mathbf{z}_{p}, k \mid \mathbf{x}\right) \| p\left(\mathbf{z}_{p}, k\right)\right)\\
					& -D_{\mathrm{KL}}\left(q_{\phi}\left(\mathbf{z}_{l} \mid \mathbf{x}\right) \| p\left(\mathbf{z}_{l}\right)\right) 
				\end{aligned}
			\end{equation}

			\subsection{Experiment Setup}\label{data}
			\subsubsection{Implementation}
			Following~\cite{chung1996generalized, grebenkov2014following,guo2024addressing}, we set the parameter $\alpha$ in EMA to 0.99 to learn global class-related information. The main experimental setup involves 20 clients collaborating in training, while the ablation study extends the analysis to 50 clients. The centralized federated learning baseline methods are evaluated within a server-supported framework, whereas the decentralized federated learning baselines are implemented under a ring topology for subsequent experimental comparisons.

			\subsubsection{Dataset and Data partition }
			The SVHN dataset, designed for digit classification, contains 600,000 32 $ \times $ 32 RGB images of printed digits extracted from Street View house numbers. For our experiments, we utilize a subset comprising 33,402 images for training and 13,068 images for testing.
			CIFAR-10 is a comprehensive image dataset comprising 10 classes, with each class containing 6,000 samples of size 32 $ \times $ 32. Similarly, CIFAR-100 is an extended version with 100 classes, where each class includes 600 samples of the same size, offering finer granularity for image classification tasks.

			\begin{figure*}[!htb]
				\vskip 0.2in
				\begin{center}
					\begin{minipage}[b]{0.4\linewidth}
						\centering
						\includegraphics[width=\linewidth]{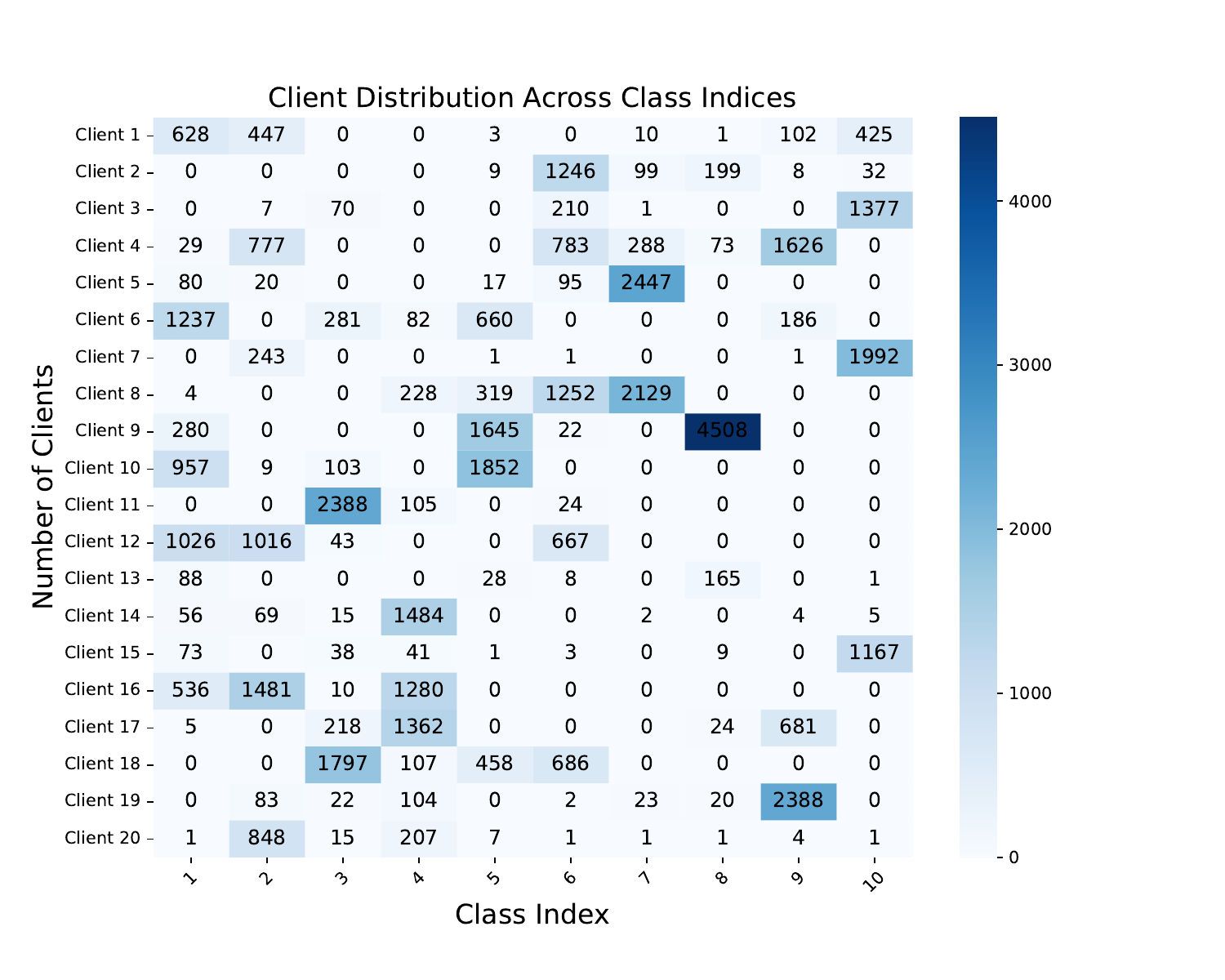}
						\subcaption{CIFAR-10 with $ \beta $ = 0.1}
						\label{fig:non-iid1}
					\end{minipage}
					\begin{minipage}[b]{0.4\linewidth}
						\centering
						\includegraphics[width=\linewidth]{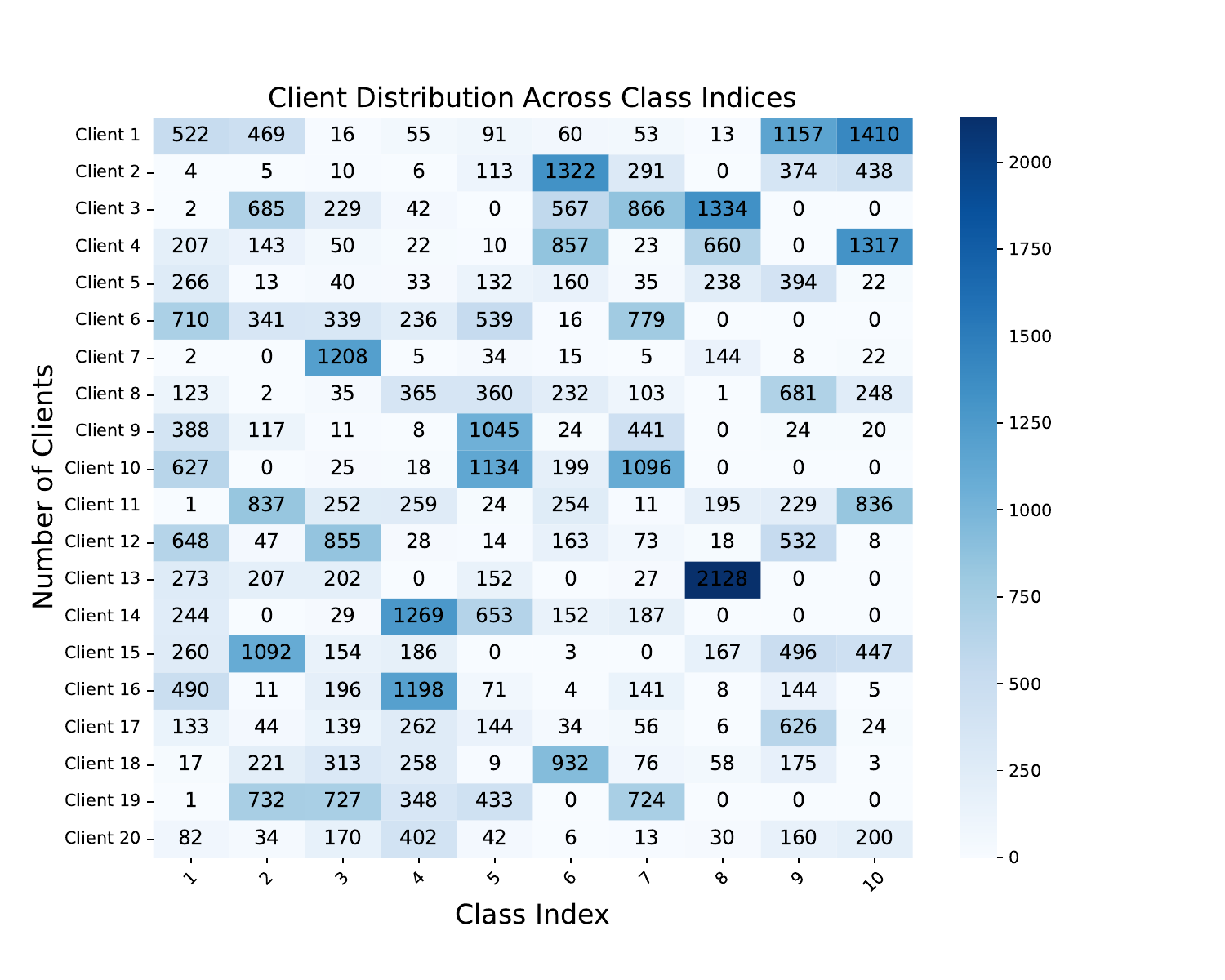}
						\subcaption{CIFAR-10 with $ \beta $ = 0.4}
						\label{fig:non-iid2}
					\end{minipage}
					
					\begin{minipage}[b]{0.4\linewidth}
						\centering
						\includegraphics[width=\linewidth]{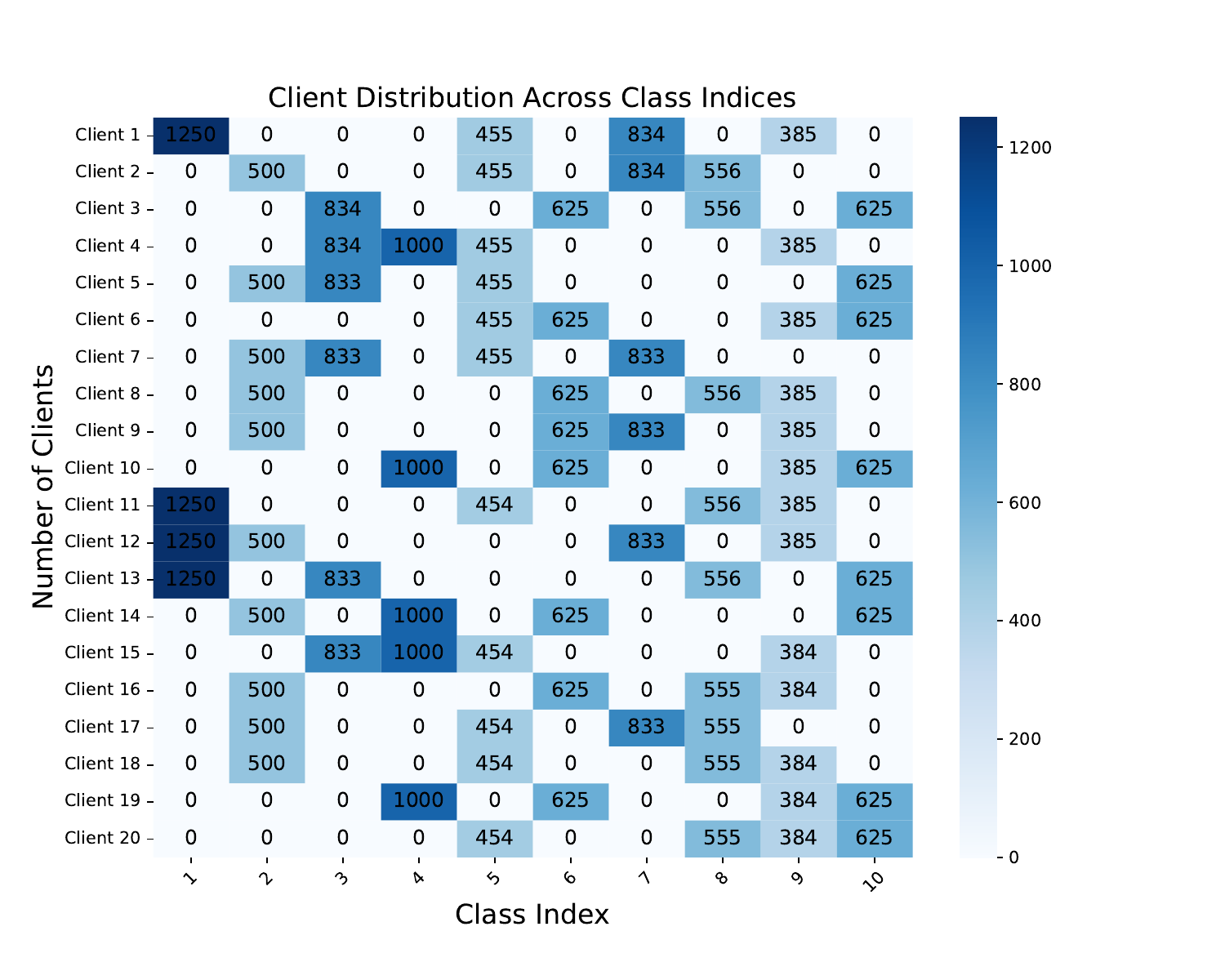}
						\subcaption{CIFAR-10 with $ s $ = 4}
						\label{fig:non-iid3}
					\end{minipage}
					\begin{minipage}[b]{0.4\linewidth}
						\centering
						\includegraphics[width=\linewidth]{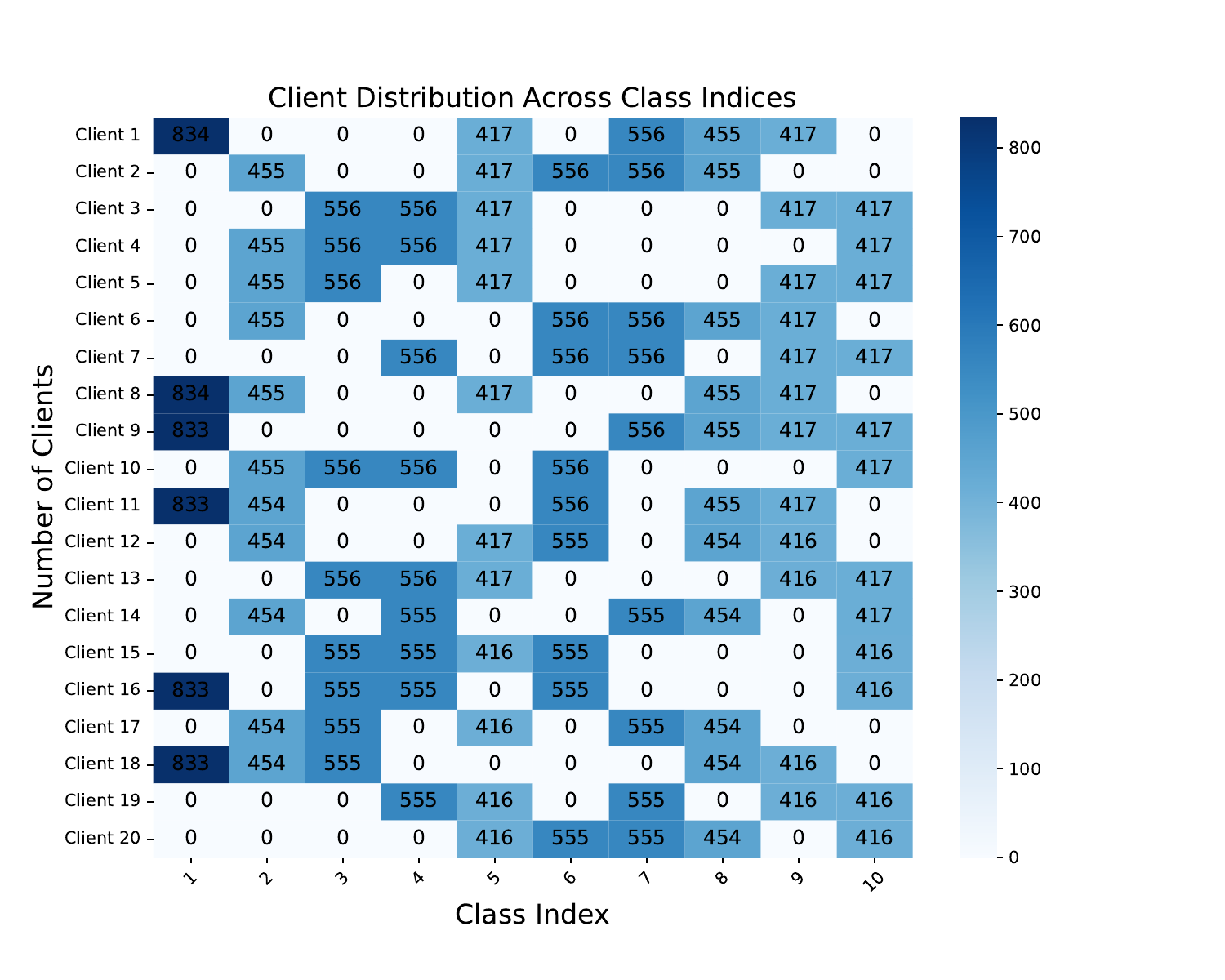}
						\subcaption{CIFAR-10 with $ s $ = 5}
						\label{fig:non-iid4}
					\end{minipage}
					\caption{The Non-IID data distribution simulated on different clients based on the CIFAR-10 dataset within the RDfl architecture.}
					\label{fig:non-iid}
				\end{center}
				\vskip -0.2in
			\end{figure*}
			Figure~\ref{fig:non-iid} illustrates the data distributions under different non-IID settings based on the CIFAR-10 dataset. Figures~\ref{fig:non-iid1} and~\ref{fig:non-iid2} show the client data distributions in the Dirichlet-based Non-IID scenario, where both the class distribution and the number of samples vary across classes. In contrast, Figures~\ref{fig:non-iid3} and~\ref{fig:non-iid4} represent the Shard-based Non-IID scenario, where each client has a distinct class distribution, but the number of samples per class remains identical. Both scenarios effectively simulate the problem of label distribution shift in data heterogeneity. Moreover, the test data shares the same class distribution as the training data but is composed of different samples, thereby modeling the feature distribution shift inherent in data heterogeneity.
			
			\subsubsection{Baselines}
			\begin{itemize}
				\item \textbf{Local} is the direct solution to the personalized federated learning problem. Each client only performs SGD on their own data. For the sake of consistency, we take 5 epochs of local training as one communication round.
				\item \textbf{FedRep}~\cite{collins2021exploiting} is a classic personalized federated learning method. It achieves personalized model training by sharing part of the model with the server during communication and training a personalized head locally. In our setup, the number of locally shared model training epochs is set to 4, and the number of personalization epochs is set to 1. 
				\item \textbf{FedNova}~\cite{wang2020tackling} employs a normalized averaging approach to eliminate objective inconsistency while maintaining fast error convergence. This method ensures that models trained on Non-IID data reduce objective inconsistencies, thereby improving the generalization performance of the global model.   
				\item \textbf{FedBN}~\cite{li2021fedbn} is a federated learning method based on the personalization of Batch Normalization (BN). Each client retains its personal BN layer statistics, including mean and variance, while other model parameters, such as weights and biases of convolutional and fully connected layers, are aggregated and shared among clients.  
				\item \textbf{FedFed}~\cite{yang2024fedfed} introduces a data-driven approach that divides the underlying data into performance-sensitive features (which contribute significantly to model performance) and performance-robust features (which have limited impact on model performance). Performance-sensitive features are globally shared to mitigate data heterogeneity, while performance-robust features are retained locally, facilitating personalized private models.  
				\item \textbf{DFedPGP}~\cite{liu2024decentralized} is a state-of-the-art personalized distributed federated learning (DFL) method. It personalizes the linear classifier of modern deep models to tailor local solutions and learns consensus representations in a fully decentralized manner. Clients share gradients only with a subset of neighbors based on a directed and asymmetric topology, ensuring resource efficiency and enabling flexible choices for better convergence.  
				\item \textbf{Fedcvae} is a comparative method we propose based on Conditional Variational Autoencoders (CVAE)~\cite{sohn2015learning} and a distributed federated learning architecture. Each client uses its private dataset trains a pretrained model \( g_\varphi(\cdot) \) to obtain the prior distribution and the CVAE model \( f_w(\cdot) \) and a classifier \( C_\omega(\cdot) \) until convergence. The CVAE consists of an encoder \( E_\phi(\cdot) \), a decoder \( D_\theta(\cdot) \) with parameters denoted as \( w =[\phi, \theta] \). Collaborative learning among clients is achieved by using the pretrained model as the shared interaction information.
				Previous research on CVAE has explored its application in defense against malicious clients~\cite{wen2020unified,gu2021detecting}. In one-shot federated learning~\cite{heinbaugh2023data}, an ensemble dataset is constructed at the server to train a server-side classifier. In federated learning frameworks~\cite{kasturi2022communication} based on VAE, client-generated data is aggregated at the server to train a global model. However, this approach is different from our learning goals and the decentralized learning scenario we are focusing on.
				
				\item \textbf{DisPFL}~\cite{dai2022dispfl} is a classical personalized federated learning method in distributed scenarios. It uses personalized sparse masks to customize edge-local sparse models. During point-to-point communication, each local model maintains a fixed number of active parameters throughout the local training process, reducing communication costs.
			\end{itemize}
			\subsection{Additional Experimental Results}\label{aer}
\begin{figure*}[!htb]
				\vskip 0.2in
				\begin{center}
					\begin{minipage}[b]{0.24\linewidth}
						\centering
						\includegraphics[width=\linewidth]{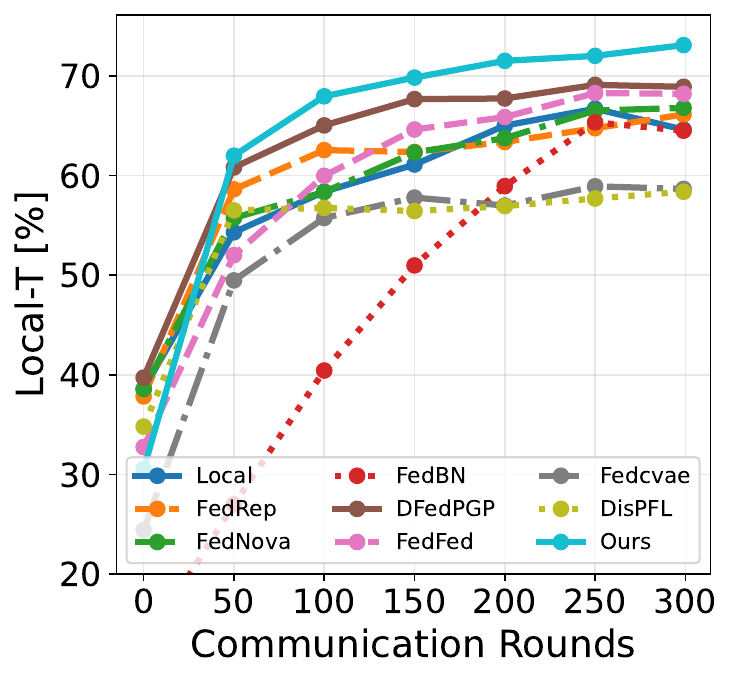}
						\subcaption{CIFAR-100 with $ \beta $ = 0.1}
						\label{fig:subplot1}
					\end{minipage}
					\begin{minipage}[b]{0.24\linewidth}
						\centering
						\includegraphics[width=\linewidth]{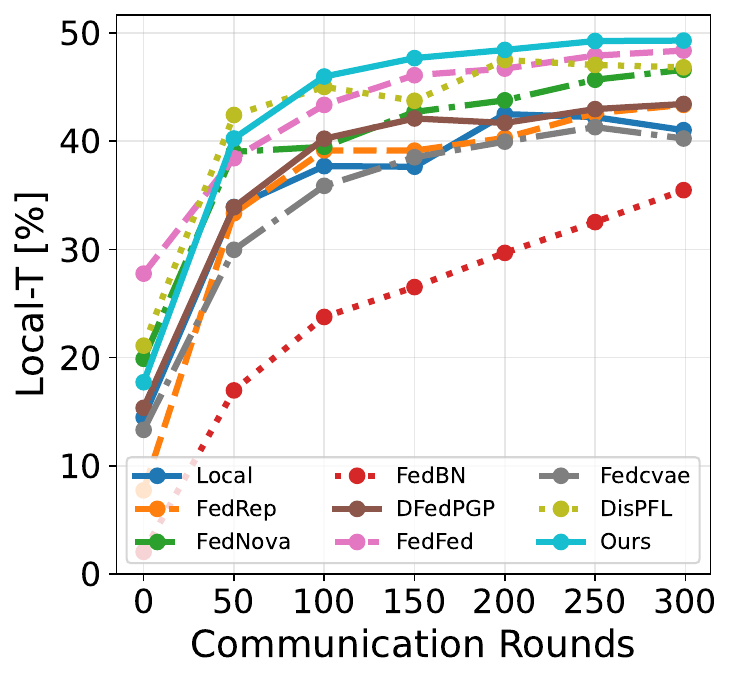}
						\subcaption{CIFAR-100 with $ \beta $ = 0.4}
						\label{fig:subplot2}
					\end{minipage}
					\begin{minipage}[b]{0.24\linewidth}
						\centering
						\includegraphics[width=\linewidth]{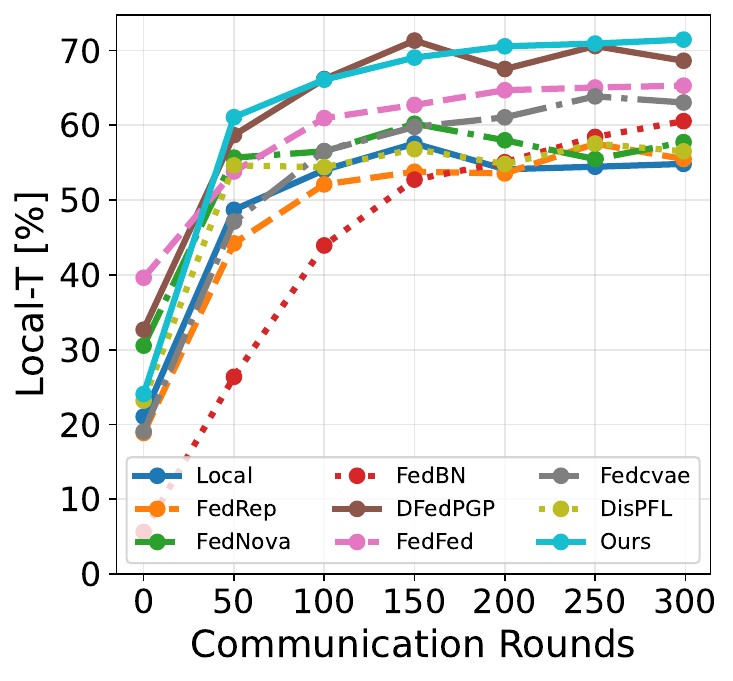}
						\subcaption{CIFAR-100 with $ s $ = 20}
						\label{fig:subplot3}
					\end{minipage}
					\begin{minipage}[b]{0.24\linewidth}
						\centering
						\includegraphics[width=\linewidth]{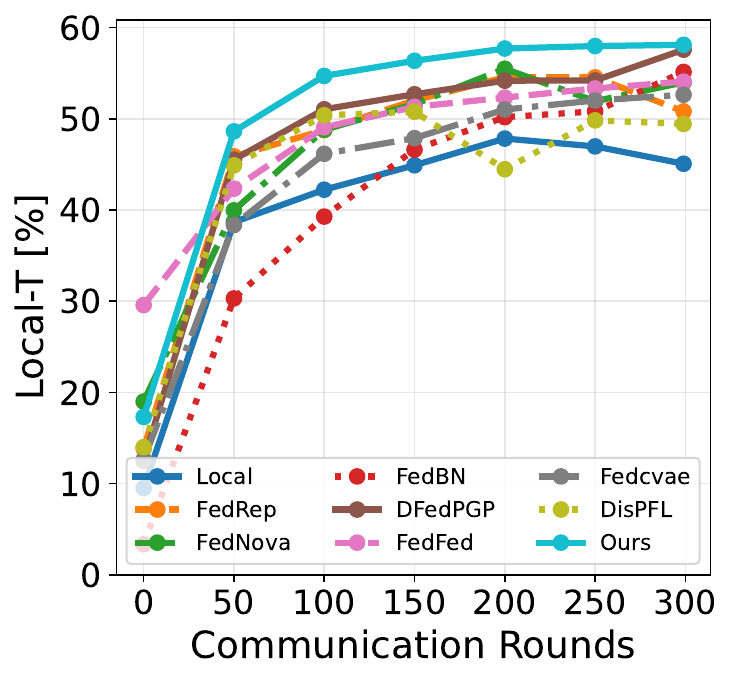}
						\subcaption{CIFAR-100 with $ s $ = 30}
						\label{fig:subplot4}
					\end{minipage}
					\caption{Comparison of personalized performance curves for different methods under various Non-IID settings on the CIFAR-100 dataset.}
					\label{fig:cifar100}
				\end{center}
			\end{figure*}

			\textbf{Convergence analysis.}
			In Figure~\ref{fig:cifar100}, we present a comparative analysis of personalized performance across various methods on the CIFAR-100 dataset under different Non-IID settings. Similar to the trends observed in Figure~\ref{CON_CIFAR10} for the CIFAR-10 dataset, our method demonstrates a smooth convergence curve and outperforms other approaches in most cases. In the CIFAR-100 setting with \(s=20\), although the DFedPGP method achieves higher performance in some rounds, it exhibits more fluctuations. Particularly, in the CIFAR-100 setting with \(\alpha = 0.1\) and \(s = 30\), our method achieves higher accuracy with fewer communication rounds, highlighting its superior convergence speed. The results demonstrate that our proposed method, consistently outperforms other baseline approaches, such as FedRep, FedNova, FedBN, and DisPFL, in terms of Local-T, which indicates the model's ability to personalize effectively across clients. When considering convergence behavior, the proposed method also demonstrates faster convergence compared to the other methods. The model reaches higher Local-T with fewer communication rounds, highlighting its efficiency in both convergence speed and resource utilization.
			
\begin{figure}[htb]
				\begin{center}
					\centerline{\includegraphics[width=0.5\linewidth]{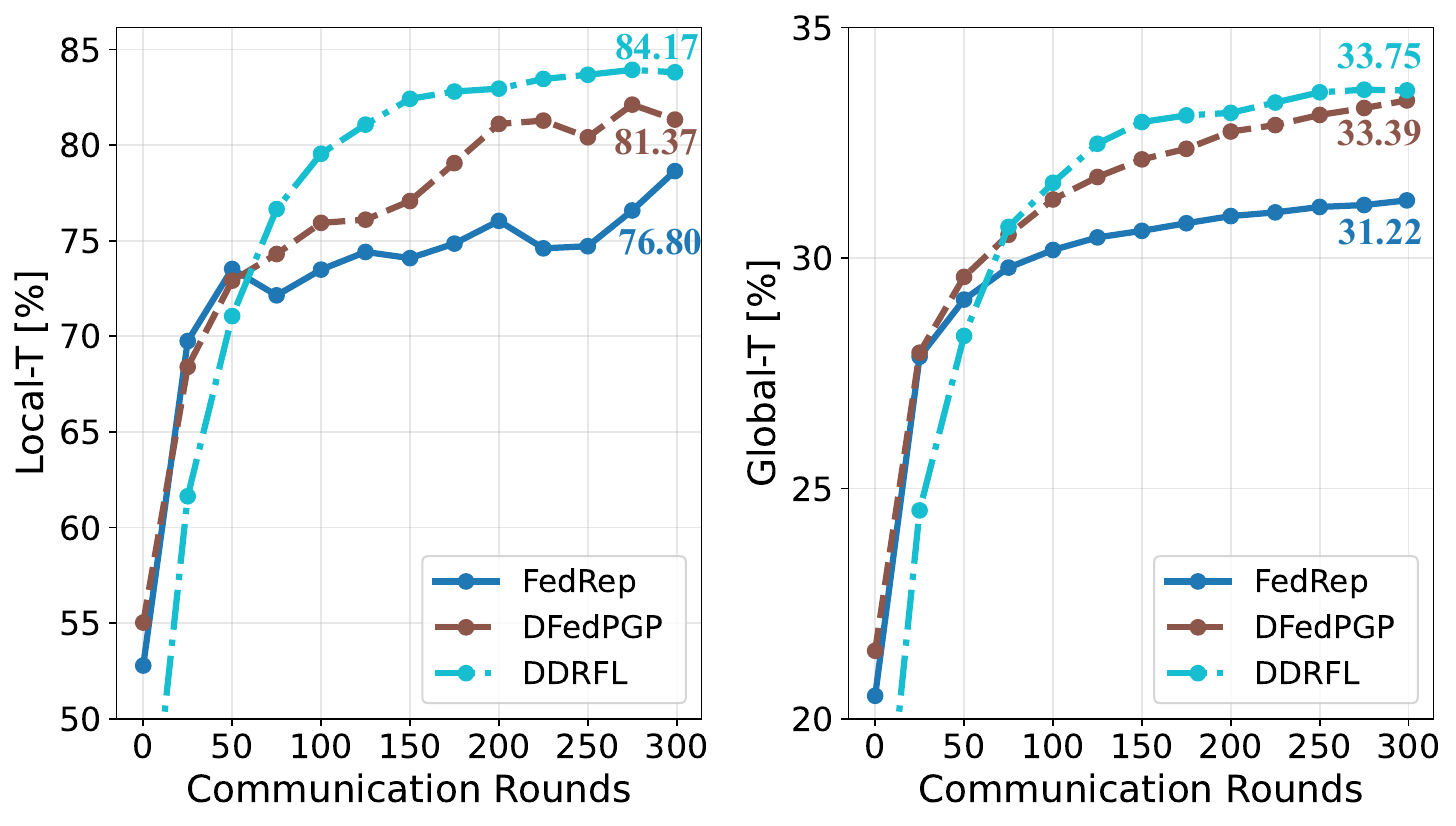}}
					\caption{Comparison of Local-T and Global-T curves for different personalized methods on CIFAR-10 with $s$ = 4 across 50 clients.}
					\label{fig:50clients}
				\end{center}
			\end{figure}

			\textbf{Ablation study.} To further validate the applicability of DRDFL across different client scales, we perform collaborative learning with 50 clients and compare it to the FedRep method in CFL and the DFedPGP method in DFL, as shown in Figure~\ref{fig:50clients}. DRDFL achieves significant improvements in both convergence speed and performance on the Local-T and Global-T metrics. The numbers in the figure represent the average values of the last 10 rounds, with DRDFL outperforming FedRep by 7.37\% on Local-T and slightly outperforming it by 2.53\% on Global-T.
			
			\begin{figure}[!htb]
				\begin{center}
					\centerline{\includegraphics[width=0.8\linewidth]{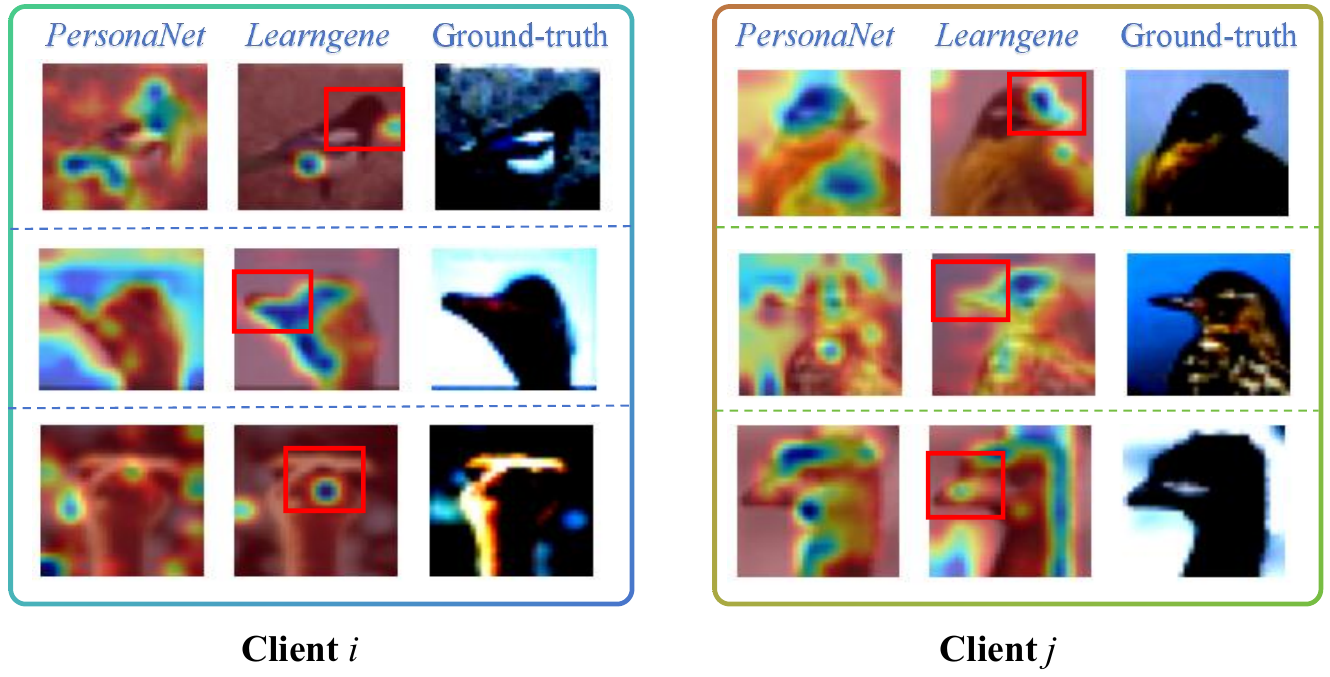}}
					\caption{Visualization of \textbf{bird} class samples from CIFAR-10 for different clients. The \textit{PersonaNet} column presents the Grad-CAM~\cite{selvaraju2017grad} outputs generated by the \textit{PersonaNet} module, while the \textit{Learngene} column illustrates the shared general representations captured by the \textit{Learngene} module. The regions highlighted with red boxes emphasize the shared attention patterns.}
					\label{fig:cam}
				\end{center}
			\end{figure}
			
			\textbf{Grad-CAM Visualization of \textit{Learngene} and \textit{PersonaNet} Representations.} To further validate the framework’s capability in capturing both generalized and personalized information, we conduct Grad-CAM visualizations of the \textit{PersonaNet} and Learngene modules on ``bird" category image samples from different clients, as illustrated in Figure~\ref{fig:cam}. The ``Ground-truth" column provides the original input images for reference and comparative analysis. It is clear that the activation maps generated by \textit{PersonaNet} reflect client-specific attention regions, highlighting personalized patterns learned by each client model. In contrast, the \textit{Learngene} module consistently focuses on semantically meaningful and discriminative regions across clients—such as the bird's head and beak. These shared attention patterns are marked with red bounding boxes, indicating the common knowledge captured by Learngene across different clients.
			This observation confirms that \textit{Learngene} is capable of learning generalized representations that maintain consistent focus on task-relevant semantic regions, regardless of the client-specific variations. Such representations are particularly valuable for initializing models of newly joined clients in federated learning, as they provide strong prior knowledge to facilitate fast convergence toward effective personalized models.
			
				\begin{figure}[!htb]
				\begin{center}
					\centerline{\includegraphics[width=0.8\linewidth]{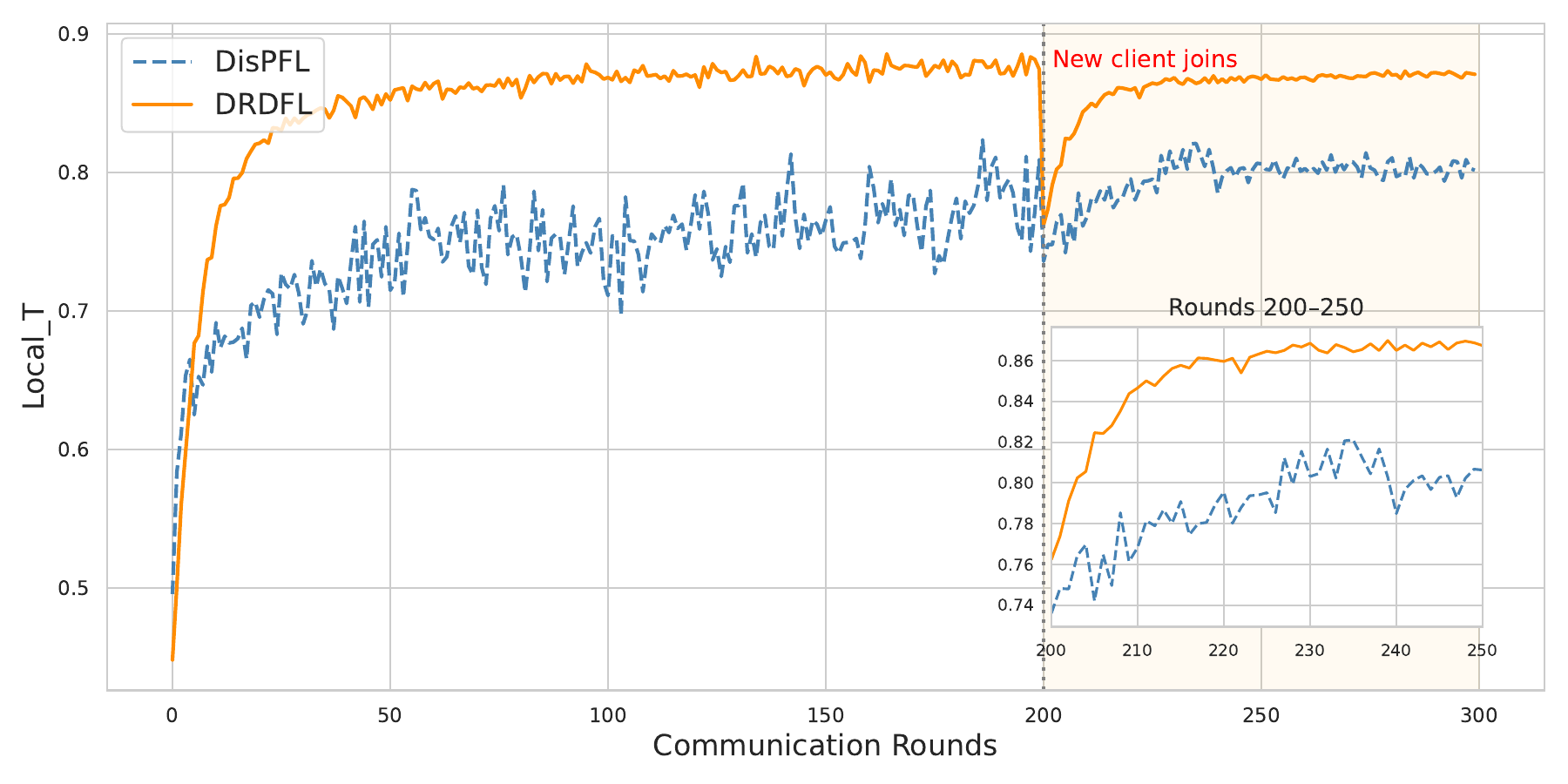}}
					\caption{Visualization the average performance of new clients joining the ring-topology federated learning system on CIFAR-10. DRDFL provides strong initialization using the optimized \textit{Learngene} and global priors, leading to rapid convergence.}
					\label{fig:new_client}
				\end{center}
			\end{figure}
			\textbf{Adaptation to Newly Joined Clients in Ring-Topology Federated Learning.} 
			The ring topology offers excellent scalability, enabling new clients to dynamically join the federated learning system. However, this flexibility also introduces a new challenge: how to effectively initialize models for newly added clients. The \textit{Learngene} module we designed, which encapsulates generalized knowledge-capturing transferable and generalizable representations-can seamlessly adapt to unknown clients.

			We empirically validate this hypothesis through a two-stage experimental setup. In the first stage, a ring-topology federated learning system with 15 clients undergoes 200 rounds of collaborative training to ensure convergence. In the second stage, five new clients with previously unseen data distributions are introduced into the system. The average performance of two methods on participated clients is illustrated in Fig.~\ref{fig:new_client}.  As shown, our proposed DRDFL method leverages the optimized \textit{Learngene} and global Gaussian information to provide strong model initialization for the new clients, significantly accelerating their convergence. In contrast, DisPFL maintains a fixed number of active parameters and exhibits unstable performance when adapting to new clients during collaborative training.
			
			\begin{figure}[!htb]
				\begin{center}
					\centerline{\includegraphics[width=1\linewidth]{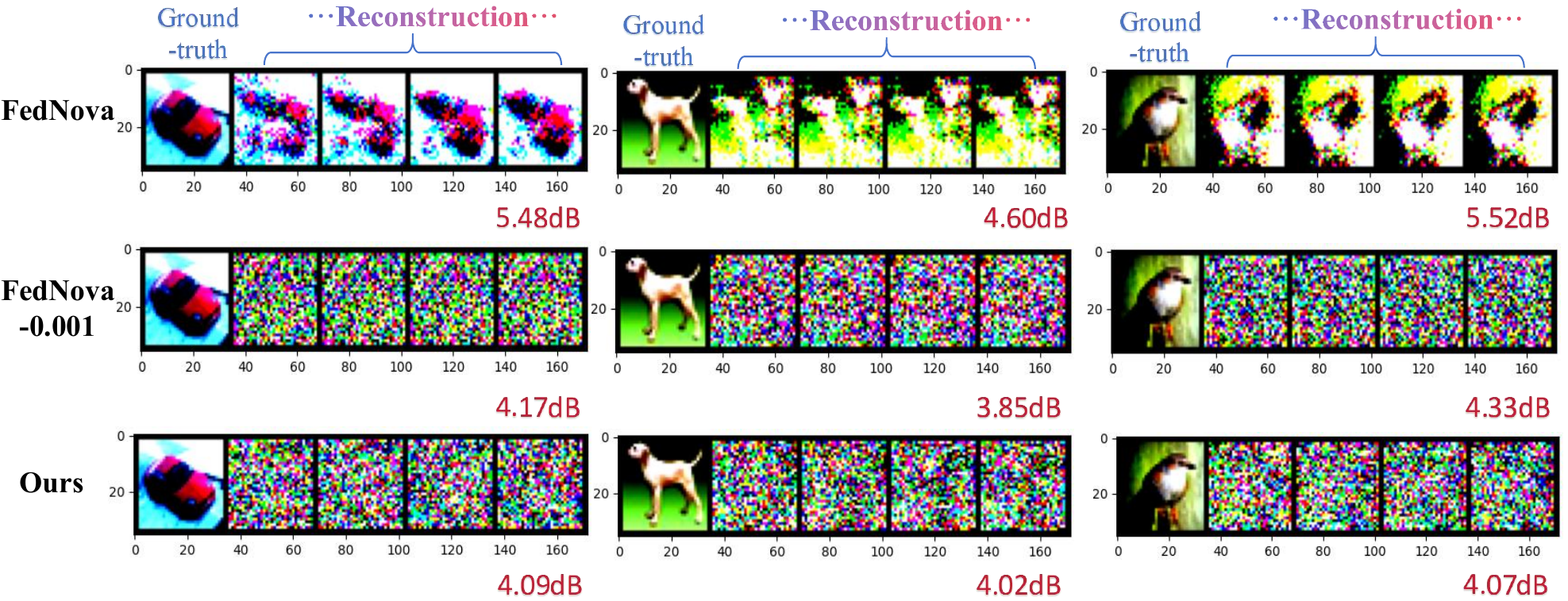}}
					\caption{Visualization of the image reconstruction process based on leaked information, along with the difference values (Peak Signal to Noise Ratio) between the final reconstructed images after 200 iterations and the original images.}
					\label{fig:psnr}
				\end{center}
			\end{figure}
			
			\textbf{Robustness to Gradient-based Attack.}
			Gradient inversion attacks, such as Deep Leakage from Gradients (DLG)~\cite{zhu2019deep}, pose a significant threat to federated learning frameworks that transmit raw model gradients. DLG works by iteratively optimizing dummy inputs to match the shared gradients, gradually reconstructing the original training data through repeated loss and gradient computations. 
			In contrast, our framework fundamentally prevents such leakage by not exposing raw gradients or instance-level representations. Instead, we transmit only the \textit{Learngene} module and class-wise Gaussian distributions produced by the \textit{PersonaNet} module. These distributions summarize task-relevant characteristics without retaining any recoverable instance-specific detail, making loss rehearsal infeasible for the attacker.
			
			To validate this, we conduct reconstruction experiments using CIFAR-10 and CIFAR-100, comparing FedNova, FedNova with added Gaussian noise, and our method. Each attack is performed for 200 optimization steps. As shown in Figure~\ref{fig:psnr}, the attacker successfully recovers recognizable images from FedNova. However, when applying the same attack to our method, the reconstructed results are visually unrecognizable and yield significantly lower PSNR scores.
			This demonstrates that the Learngene module effectively eliminates reconstructable gradients by transmitting only distributional representations. Consequently, our method enables secure and generalizable model initialization in federated settings without risking data leakage.
			\subsection{Analysis}
			Before analyzing the convergence of DRDFL, we first introduce additional notation. Let \( t \) denote the communication round among clients, and let \( e \in \{0, 1, \ldots, E\} \) represent the local training epoch or iteration within each client. The iteration index \( tE + e \) corresponds to the \( e \)-th local update in the \((t + 1)\)-th communication round. Specifically, \( tE + 0 \) refers to the point at which, in the \((t + 1)\)-th round, clients receive the learning gene generated in the \( t \)-th round prior to commencing local training. Conversely, \( tE + E \) denotes the final iteration of local training, marking the completion of local updates in the \((t + 1)\)-th round. For simplicity, we assume that all models adopt a uniform learning rate \( \eta \).
			
			\textbf{Assumption 1. Lipschitz Smoothness.} Gradients of client $m$'s local complete model $\boldsymbol{w}_{m}$ are L1-Lipschitz smooth~\cite{tan2022fedproto,yi2024pfedmoe},
			\begin{equation}
				\begin{array}{c}
					\left\|\nabla \mathcal{L}_{m}^{t_{1}}\left(\boldsymbol{w}_{m}^{t_{1}} ; \boldsymbol{x}, y\right)-\nabla \mathcal{L}_{m}^{t_{2}}\left(\boldsymbol{w}_{m}^{t_{2}} ; \boldsymbol{x}, y\right)\right\|\le L_{1}\left\|\boldsymbol{w}_{m}^{t_{1}}-\boldsymbol{w}_{m}^{t_{2}}\right\|, \\
					\forall t_{1}, t_{2}>0, m \in\{0,1, \ldots, M-1\},(\boldsymbol{x}, y) \in D_{m}.
				\end{array}
			\end{equation}
			
			The above formulation can be further expressed as:\\
			\begin{equation}
				\mathcal{L}_{m}^{t_{1}}-\mathcal{L}_{m}^{t_{2}} \le \left \langle\nabla  \mathcal{L}_{m}^{t_{2}},\left(\boldsymbol{w}_{m}^{t_{1}}-\boldsymbol{w}_{m}^{t_{2}}\right)\right\rangle+\frac{L_{1}}{2}\left\|\boldsymbol{w}_{m}^{t_{1}}-\boldsymbol{w}_{m}^{t_{2}}\right\|_{2}^{2}.
			\end{equation}
			
			\textbf{Assumption 2. Unbiased Gradient and Bounded Variance.} The client $m$'s random gradient $g_{\boldsymbol{w}, m}^{t}=\nabla \mathcal{L}_{m}^{t}\left(\boldsymbol{w}_{m}^{t} ; \xi_{m}^{t}\right)(\xi$ is a batch of local data) is unbiased,
			\begin{equation}\begin{array}{c}
					\mathbb{E}_{\xi_{m}^{t} \subseteq D_{m}}\left[g_{\boldsymbol{w}, m}^{t}\right]=\nabla \mathcal{L}_{m}^{t}\left(\boldsymbol{w}_{m}^{t}\right),
			\end{array}\end{equation}
			and the variance of  $g_{\boldsymbol{w}, m}^{t}$ is bounded by:
			\begin{equation}\begin{array}{c}
					\mathbb{E}_{\xi_{m}^{t} \subseteq D_{m}}\left[\left\|\nabla \mathcal{L}_{m}^{t}\left(\boldsymbol{w}_{m}^{t} ; \xi_{m}^{t}\right)-\nabla \mathcal{L}_{m}^{t}\left(\boldsymbol{w}_{m}^{t}\right)\right\|_{2}^{2}\right] \le \sigma^{2}.
			\end{array}\end{equation}
			
			\textbf{Assumption 3. Bounded Parameter Variation.} The parameter variations of the homogeneous small \textit{Learngene} $\phi_{m}^{t}$ and $\tilde{\phi}^{t}$ before and after aggregation is bounded as:
			
			\begin{equation}\left\|\tilde{\phi}^{t}-\phi_{m}^{t}\right\|_{2}^{2} \leq \delta^{2}.\end{equation}
			
			Based on the above assumptions, we can derive the following Lemma and Theorem. 
			
			\textbf{Lemma 1. Local Training.} Based on Assumptions 1 and 2, during  $\{0,1, \ldots, E\}$  local iterations of the ($t + 1$)-th FL training round, the loss of an arbitrary client's local model is bounded by:
			\begin{equation}\begin{aligned}
					\mathrm{E}\left[\mathcal{L}_{(t+1) E}\right] & \leq \mathcal{L}_{t E+0}+\left(\frac{L_{1} \eta^{2}}{2}-\eta \right) \sum_{e=0}^{E-1}\left\|\nabla \mathcal{L}_{t E+e}\right\|_{2}^{2}+\frac{L_{1} \eta^{2}\sigma^{2}}{2}
				\end{aligned}.\end{equation}
			
			\textbf{Lemma 2. \textit{Learngene} Aggregation.} Given Assumptions 2 and 3, after the ($t+1$) -th local training round, the loss of client before and after aggregating local homogeneous small \textit{Learngene} is bounded by
			
			\begin{equation}\mathbb{E}\left[\mathcal{L}_{(t+1) E+0}\right] \leq \mathbb{E}\left[\mathcal{L}_{t E+1}\right]+\eta \delta^{2}.\end{equation}
			
			\textbf{Theorem 1. One Complete Round of FL.} Based on Lemma 1 and Lemma 2, we get
			
			\begin{equation}\mathbb{E}\left[\mathcal{L}_{(t+1) E+0}\right] \leq \mathcal{L}_{t E+0}+\left(\frac{L_{1} \eta^{2}}{2}-\eta\right) \sum_{e=0}^{E}\left\|\nabla \mathcal{L}_{t E+e}\right\|_{2}^{2}+\frac{L_{1} E \eta^{2} \sigma^{2}}{2}+\eta \delta^{2}.\end{equation}
			
			\textbf{Theorem 2. Non-convex Convergence Rate of DRDFL.} Based on Theorem 1, for any client and an arbitrary constant $\epsilon>0$, the following holds true:
			
			\begin{equation}\begin{aligned}
					\frac{1}{T} \sum_{t=0}^{T-1} \sum_{e=0}^{E-1}\left\|\nabla \mathcal{L}_{t E+e}\right\|_{2}^{2} & \leq \frac{\frac{1}{T} \sum_{t=0}^{T-1}\left[\mathcal{L}_{t E+0}-\mathbb{E}\left[\mathcal{L}_{(t+1) E+0}\right]\right]+\frac{L_{1} E \eta^{2} \sigma^{2}}{2}+\eta \delta^{2}}{\eta-\frac{L_{1} \eta^{2}}{2}}<\epsilon, \\
					\text { s.t. } \eta & <\frac{2\left(\epsilon-\delta^{2}\right)}{L_{1}\left(\epsilon+E \sigma^{2}\right)} .
			\end{aligned}\end{equation}
			
			Therefore, we conclude that any client's local model can converge at a non-convex rate $\epsilon \sim \mathcal{O}\left(\frac{1}{T}\right)$ under DRDFL.
			
			\subsection{THEORETICAL PROOFS}
			
			\subsubsection{Proof for Lemma 1}
				An arbitrary client $m$'s local model $\boldsymbol{w}$ can be updated by $\boldsymbol{w}_{t+1} = \boldsymbol{w}_t - \eta g_{\boldsymbol{w}_t}$ in the $(t+1)$-th round, and following Assumption 1, we can obtain:
			
			\begin{equation} \begin{aligned}	
					\mathcal{L}_{t+1} &\leq \mathcal{L}_{t} + \langle \nabla \mathcal{L}_{tE+0}, (\boldsymbol{w}_{tE+1} - \boldsymbol{w}_{tE+0}) \rangle + \frac{L_1}{2} \|\boldsymbol{w}_{tE+1} - \boldsymbol{w}_{tE+0} \|^2 \\
					&= \mathcal{L}_{tE+0} - \eta \langle \nabla \mathcal{L}_{tE+0}, g_{\boldsymbol{w},tE+0} \rangle + \frac{L_1\eta^2}{2} \| g_{\boldsymbol{w},tE+0} \|^2.
			\end{aligned}\end{equation} 			
			Taking the expectation of both sides of the inequality concerning the random variable $\xi_{tE+0}$, we obtain:
			
			\begin{equation} 
				\begin{aligned}	
					\mathbb{E}\big[\mathcal{L}_{tE+1}\big] &\leq \mathcal{L}_{tE+0} - \eta\mathbb{E}\big[\langle\nabla\mathcal{L}_{tE+0}, g_{\boldsymbol{w},tE+0}\rangle\big] + \frac{L_1\eta^{2}}{2}\mathbb{E}\big[\|g_{\boldsymbol{w},tE+0}\|_{2}^{2}\big] \\
					&\overset{(a)}{\leq} \mathcal{L}_{tE+0} - \eta\|\nabla\mathcal{L}_{tE+0}\|_{2}^{2} + \frac{L_1\eta^{2}}{2}\mathbb{E}\big[\|g_{\boldsymbol{w},tE+0}\|_{2}^{2}\big] \\
					&\overset{(b)}{\leq} \mathcal{L}_{tE+0} - \eta\|\nabla\mathcal{L}_{tE+0}\|_{2}^{2} + \frac{L_1\eta^{2}}{2}\left(\mathbb{E}\big[\|g_{\boldsymbol{w},tE+0}\|_{2}^{2} + \mathrm{Var}(g_{\boldsymbol{w},tE+0})\big]\right)\\
					&\overset{(c)}{\leq} \mathcal{L}_{tE+0} - \eta\|\nabla\mathcal{L}_{tE+0}\|_{2}^{2} + \frac{L_1\eta^{2}}{2}\left(\|\nabla\mathcal{L}_{tE+0}\|_{2}^{2} + \mathrm{Var}(g_{\boldsymbol{w},tE+0})\right) \\
					&\overset{(d)}{\leq} \mathcal{L}_{tE+0} - \eta\|\nabla\mathcal{L}_{tE+0}\|_{2}^{2} + \frac{L_1\eta^{2}}{2}\left(\|\nabla\mathcal{L}_{tE+0}\|_{2}^{2} + \sigma^{2}\right) \\
					&= \mathcal{L}_{tE+0} + \left(\frac{L_1\eta^{2}}{2} - \eta\right)\|\nabla\mathcal{L}_{tE+0}\|_{2}^{2} + \frac{L_1\eta^{2}\sigma^{2}}{2}, 
				\end{aligned}
			\end{equation} 			
			where (a), (c), (d) follow Assumption 2. (b) follows $\mathrm{Var}(x) = \mathbb{E}[x^{2}] - \langle\mathbb{E}[x]^{2}\rangle$.
			
			Taking the expectation of both sides of the inequality for the model $\boldsymbol{w}$ over $E$ iterations, we obtain
			
			\begin{equation}
				\mathbb{E}\big[\mathcal{L}_{tE+1}\big] \leq \mathcal{L}_{tE+0} + \left(\frac{L_1\eta^{2}}{2} - \eta\right)\sum_{i=1}^{E}\|\nabla\mathcal{L}_{tE+e}\|_{2}^{2} + \frac{L_{1}E\eta^{2}\sigma^{2}}{2}. 
			\end{equation}
			
			\subsubsection{Proof for Lemma 2}
			\begin{equation} 
				\begin{aligned}	
					\mathcal{L}_{(t+1)E+0} &= \mathcal{L}_{(t+1)E} + \mathcal{L}_{(t+1)E+0} - \mathcal{L}_{(t+1)E} \\
					&\overset{(a)}{\approx } \mathcal{L}_{(t+1)E} + \eta \| \phi_{(t+1)E+0} - \phi_{(t+1)E} \|^2_2 \\
					&\overset{(b)}{\le } \mathcal{L}_{(t+1)E} + \eta \delta^2, 
				\end{aligned}
			\end{equation} 			
			
			where (a): we can use the gradient of parameter variations to approximate the loss variations, i.e., $\Delta\mathcal{L}\approx\eta\cdot\|\Delta\phi\|_{2}^{2}$. (b) follows Assumption 3. Taking the expectation of both sides of the inequality to the random variable $\xi$, we obtain
			
			\begin{equation}
				\mathbb{E}\left[\mathcal{L}_{(t+1)E+0}\right]\leq\mathbb{E}\left[ \mathcal{L}_{tE+1}\right]+\eta\delta^{2}.
			\end{equation}
			
			\subsubsection{Proof for Theorem 1}
			
			Substituting Lemma 1 into the right side of Lemma 2's inequality, we obtain
			
			\begin{equation}
				\label{Theorem 1}
				\mathbb{E}[\mathcal{L}_{(t+1)E+0}]\leq\mathcal{L}_{tE+0}+(\frac{L_{1}\eta^{2}}{2}-\eta)\sum_{e=0}^{E}\|\nabla\mathcal{L}_{tE+e}\|_{2}^{2}+\frac{L_{1}E\eta^{2}\sigma^{2}}{2}+\eta\delta^{2}.
			\end{equation}
			
			\subsubsection{Proof for Theorem 2}
			
			Interchanging the left and right sides of Eq.~\ref{Theorem 1}, we obtain:
			
			\begin{equation}\begin{aligned}
					\sum_{e=0}^{E} \|\nabla \mathcal{L}_{tE+e}\|^2 _2
					&\leq \frac{\mathcal{L}_{tE+0} - \mathbb{E}[\mathcal{L}_{(t+1)E+0}] + \frac{L_1E\eta^2 \sigma^2}{2} + \eta \delta^2}{\eta - \frac{L_1\eta^2}{2}}.
			\end{aligned}\end{equation} 			
			
			Taking expectation over rounds $t = [0, T-1]$:
			
			\begin{equation} 				
				\begin{aligned}
					\frac{1}{T} \sum_{t=0}^{T-1} \sum_{e=0}^{E-1} \|\nabla \mathcal{L}_{tE+e}\|^2_2 
					&\leq \frac{\frac{1}{T} \sum_{t=0}^{T-1} [\mathcal{L}_{tE+0} - \mathbb{E}[\mathcal{L}_{(t+1)E+0}]] + \frac{L_1E\eta^2 \sigma^2}{2} + \eta \delta^2}{\eta - \frac{L_1\eta^2}{2}}.
			\end{aligned}\end{equation} 			
			
			Let $\Delta = \mathcal{L}_{t=0} - \mathcal{L}^* > 0$, then $\sum_{t=0}^{T-1} [\mathcal{L}_{tE+0} - \mathbb{E}[\mathcal{L}_{(t+1)E+0}]] \leq \Delta$, we get:
			
			\begin{equation} \begin{aligned}	
					\label{eq31}
					\frac{1}{T} \sum_{t=0}^{T-1} \sum_{e=0}^{E-1} \|\nabla \mathcal{L}_{tE+e}\|^2_2 
					&\leq \frac{\frac{\Delta}{T}+L_1 E \eta^2 \sigma^2+\eta \delta^2}{\eta - \frac{L_1\eta^2}{2}} .
			\end{aligned}\end{equation} 			
			
			If this converges to a constant $\epsilon$, i.e.,
			
			\begin{equation} \begin{aligned}	
					\frac{\Delta}{\eta - \frac{L_1\eta^2}{2}} + \frac{L_1 E \eta^2 \sigma^2}{2 (\eta - \frac{L_1\eta^2}{2})} + \frac{\eta \delta^2}{\eta - \frac{L_1\eta^2}{2}} < \epsilon,
			\end{aligned}\end{equation} 			
			
			then
			\begin{equation}
				T > \frac{\Delta}{\epsilon\left(\eta - \frac{L_{1}\eta^{2}}{2}\right) - \frac{L_{1}E\eta^{2}\sigma^{2}}{2} - \eta\delta^{2}}.
			\end{equation}
			
			Since $T > 0$, $\Delta > 0$, we can get solving the above inequality yields: 
			\begin{equation}
				\epsilon\left(\eta - \frac{L_{1}\eta^{2}}{2}\right) - \frac{L_{1}E\eta^{2}\sigma^{2}}{2} - \eta\delta^{2} > 0.
			\end{equation}
			
			After solving the above inequality, we can obtain: 
			\begin{equation}
				\eta < \frac{2(\epsilon - \delta^{2})}{L_{1}(\epsilon + {E}\sigma^{2})}.
			\end{equation}
			
			Since $\epsilon$, $L_1$, $\sigma^{2}$, $\delta^{2}$ are all constants greater than $0$, $\eta$ has solutions. Therefore, when the learning rate $g$ satisfies the above condition, any client's local complete heterogeneous model can converge. Notice that the learning rate of the local complete model involves $\{\eta_{\psi}, \eta_{\phi}, \eta_{\theta},\eta{\omega}\}$, so it's crucial to set reasonable them to ensure model convergence. Since all terms on the right side of Eq.~\ref{eq31} except for $\Delta/T$ are constants, $\Delta$ is also a constant, DRDFL's non-convex convergence rate is $\epsilon \sim \mathcal{O}\left(\frac{1}{T}\right)$.

		\bibliographystyle{IEEEtran} 
		\bibliography{dfl}
		\end{document}